\pdfoutput=1

\documentclass[11pt]{article}

\usepackage{ACL2023}

\usepackage{amsmath}
\usepackage{enumitem}
\usepackage{graphicx}
\usepackage{algorithm}
\usepackage{algorithmic}
\usepackage{tablefootnote}
\usepackage{footnote}
\usepackage{multirow}
\usepackage{multicol}
\usepackage{subcaption}
\usepackage{graphicx}
\usepackage{makecell}
\usepackage{setspace}
\usepackage{booktabs}
\usepackage{amsfonts}
\usepackage{tcolorbox}
\usepackage{xcolor}
\usepackage{amsmath}
\usepackage{array}
\usepackage{colortbl}
\usepackage{times}
\usepackage{latexsym}
\usepackage{xspace}

\usepackage[T1]{fontenc}

\usepackage[utf8]{inputenc}

\usepackage{microtype}

\usepackage{inconsolata}

\newcommand{\bbr}{\mathbb{R}}
\newcommand{\para}[1]{\noindent \textbf{#1}\xspace}
\newcommand{\caln}{\mathcal{N}}

\newcommand{\model}{{{\tt LLM-Forest}}}
\newcommand{\sysn}{{\tt LLM-Forest}\xspace}

\title{
LLM-Forest: Ensemble Learning of LLMs with Graph-Augmented Prompts for Data Imputation \\
}

\author{%
Xinrui He\textsuperscript{\textnormal{1}},
Yikun Ban\textsuperscript{\textnormal{1}$^\dagger$}, 
Jiaru Zou\textsuperscript{\textnormal{1}}, 
Tianxin Wei\textsuperscript{\textnormal{1}},
Curtiss B. Cook\textsuperscript{\textnormal{2}}, 
Jingrui He\textsuperscript{\textnormal{1}$^\dagger$}\\
\textsuperscript{1}University of Illinois at Urbana-Champaign, \textsuperscript{2}Mayo Clinic Arizona\\
\texttt{\{xhe33,yikunb2,jiaruz2,twei10,jingrui\}@illinois.edu}\\ 
\texttt{cook.curtiss@mayo.edu}\\
}

\begin{document}
\maketitle

\begingroup
\renewcommand\thefootnote{}
\footnotetext{$^\dagger$ Corresponding authors.}
\endgroup

\begin{abstract}

Missing data imputation is a critical challenge in various domains, such as healthcare and finance, where data completeness is vital for accurate analysis.
Large language models (LLMs), trained on vast corpora, have shown strong potential in data generation, making them a promising tool for data imputation. However, challenges persist in designing effective prompts for a finetuning-free process and in mitigating biases and uncertainty in LLM outputs. To address these issues, we propose a novel framework, \model, which introduces a "forest" of few-shot prompt learning LLM "trees" with their outputs aggregated via confidence-based weighted voting based on LLM self-assessment, inspired by the ensemble learning (Random Forest). This framework is established on a new concept of bipartite information graphs to identify high-quality relevant neighboring entries with both feature and value granularity. Extensive experiments on 9 real-world datasets demonstrate the effectiveness and efficiency of \model. 
The implementation is available at \url{https://github.com/Xinrui17/LLM-Forest}
\end{abstract}

\section{Introduction}

Handling missing data is a fundamental challenge across various domains, including datasets such as clinical reports, financial records, product characteristics, and survey data in a tabular format \cite{sterne2009multiple, 10.1145/3460231.3474268, shwartz2022tabular,hernandez2022synthetic,10.1145/3640457.3688145}.  However, these datasets frequently suffer from missing values, which can severely impact the reliability of critical decision-making processes that depend on complete and accurate data. To address this issue, feature imputation \cite{brick1996handling,troyanskaya2001missing,jazayeri2020imputation,bernardini2023novel}, the process of estimating missing values based on observed data, plays a crucial role in maintaining data integrity and ensuring meaningful analysis.

\begin{figure}[t]
  \centering
  \includegraphics[width=0.5\textwidth]{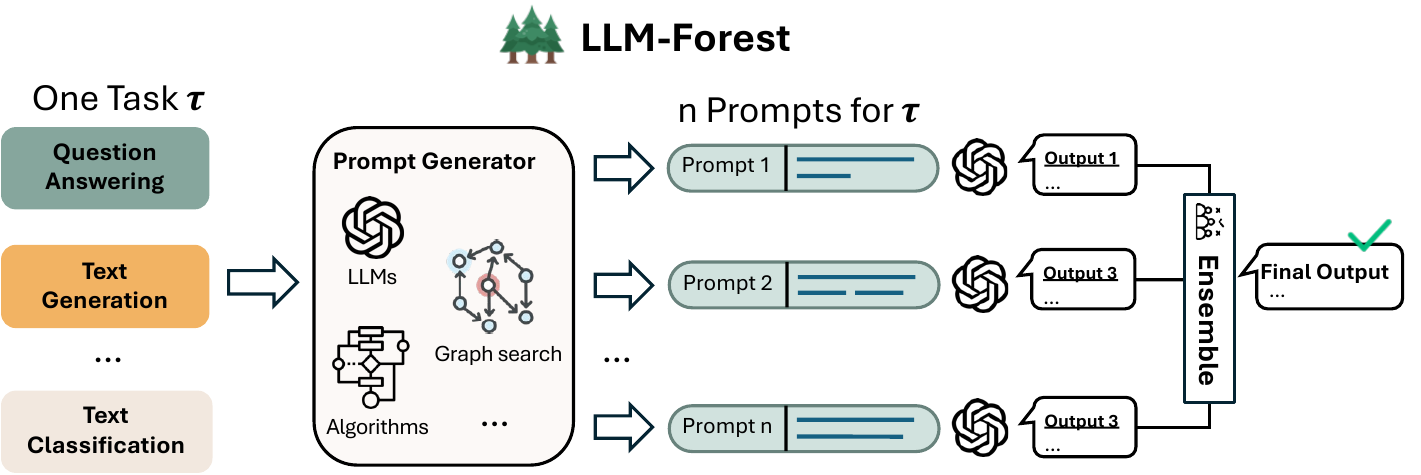}
\caption{ 
\textbf{Task-agnostic framework of \sysn: Ensemble of LLM outputs based on generated diverse prompts for a specific task.}
}
\vspace{-0.5cm}
  \label{fig:general}
\end{figure}

Common approaches for feature imputation can be broadly categorized into two groups. (1) Statistical Methods: Techniques such as mean or mode imputation and MICE \cite{little2019statistical} are simple and widely used due to their ease of implementation. While these methods are straightforward, they can introduce bias. (2) Deep Learning-Based Methods: Models like GAIN \cite{yoon2018gain}, DIFFIMPUTE \cite{mattei2019miwae}, and GRAPE \cite{you2020handling} aim to model complex data distributions and dependencies leveraging advanced neural network architectures to capture non-linear relationships within the data. However, they typically require large volumes of data and can be challenging to train, particularly in scenarios like healthcare where data are often limited, imbalanced, or subject to strict privacy regulations.

Recently, Large Language Models (LLMs) \cite{brown2020language, devlin2018bert, beltagy2019scibert,agrawal2023large,zou2025transformer,huang2025adaptive,li2025zero, 10.1145/3690624.3709196} have demonstrated strong potential in handling complex tasks by using external knowledge and contextual learning, offering a promising solution for enhancing performance on tabular data tasks \cite{gong2020tablegpt,yin2020tabert,abraham2022tablequery, thirunavukarasu2023large}. Anam \cite{nazir2023chatgpt} began exploring the adoption of ChatGPT \cite{brown2020language} as a data imputer by formulating text questions and prompting ChatGPT to respond with imputed values for each missing cell. Similarly, CLAIM \cite{hayat2024claim} introduced an LLM-based approach to generate text-specific descriptions of missing data. However, both methods require fine-tuning processes, which can be time-consuming and resource-intensive, limiting their scalability across different datasets. In contrast, in-context learning \cite{wei2022chain, min2022rethinking, gruver2024large,li2024can} has excelled in tasks such as tabular classification (e.g., CancerGPT \cite{li2024cancergpt}) and reasoning (e.g., CHAIN-OF-TABLE \cite{wang2024chain}). Yet the use of LLMs' broad knowledge and reasoning ability for data imputation, without relying on task-specific fine-tuning, remains largely unexplored.

In this paper, we explore fine-tuning-free approaches to harness the power of pre-trained LLMs for data imputation. We aim to overcome two primary challenges: (1) extracting high-quality information from observed data to align with the preferences of the target entries to effectively guide LLMs, and (2) reduce potential output bias and uncertainty in current LLMs.

To address these challenges, we propose a general and scalable framework, \sysn. It consists of multiple LLM-based trees, each guided by a distinct few-shot prompt designed to capture different perspectives of the task. These prompts can be constructed using various strategies, such as LLM-based generation, graph search, or specific algorithms. Inspired by ensemble learning like Random Forest, \sysn aggregates the predictions from all trees using confidence-based weighting, where each LLM estimates its own output confidence. As shown in Figure~\ref{fig:general}, this task-agnostic design supports a wide range of applications including question answering, text generation, and classification, offering improved robustness and reduced bias compared to using a single LLM.

In this paper, we instantiate \sysn for the task of tabular data imputation, where the goal is to fill in missing values based on observed entries. We demonstrate how the general design of \sysn can be tailored to imputation through graph-based neighbor selection. Each tree in the forest represents an LLM with a particular few-shot prompt learning process. To construct high-quality yet diverse prompts, we design a graph-based retrieval algorithm to explore highly relevant neighboring entries in preparation for constructing \sysn. This graph algorithm is based on Information-Theoretic Bipartite Graphs to quantify relevance at both the feature-level and value-level granularity. Meanwhile, it reduces the computational complexity in the graph construction process and allows for the subsequent merging process to be easily implemented in parallel. This makes it scalable for very large datasets. We then carefully design prompts to integrate the retrieved neighbors and inject domain-level guidance, maximizing the utility of pre-trained LLMs without any fine-tuning. 
Finally, we aggregate the outputs of the LLM trees by weighting each tree’s prediction according to its self-assessed confidence level, ensuring more robust and reliable results.
In summary, our \textbf{main contributions} are :
\begin{itemize}[noitemsep, left=0pt]
\item 
\textit{Stronger LLM decision-maker}: We incorporate the concept of ensemble learning, inspired by Random Forests, into the in-context learning of LLMs. By combining diverse and independent LLMs into a more powerful decision-maker, our framework enhances robustness by aggregating the outputs of multiple LLM trees, which reduces biases and the high variance associated with individual LLMs.

\item 
\textit{Information-based Graph retrieval for LLM}: We integrate information theory into the construction of bipartite information graphs to accommodate different distributions and value-level granularity, along with an efficient merging process. Random walk-based retrieval adds randomness (exploration) to each LLM tree, ensuring that the trees are less correlated with each other.

 \item
 We conduct comprehensive experiments on 9 real-world tabular datasets under various settings. The results demonstrate the effectiveness and efficiency of \sysn in imputing missing data. Our framework outperforms traditional imputation methods, highlighting its potential for practical applications in data analysis.
\end{itemize}
Since the construction of each LLM tree is independent, we can compute the predictions of LLM trees in parallel. Moreover, the graph retrieval component is efficient, enabling the framework to handle large-scale datasets by retrieving a small number of high-quality data objects for each LLM tree. This ensures that the inference time of LLM-Forest does not increase significantly in large-scale datasets.

\begin{figure*}[!th]
  \centering
  \includegraphics[width=\textwidth]{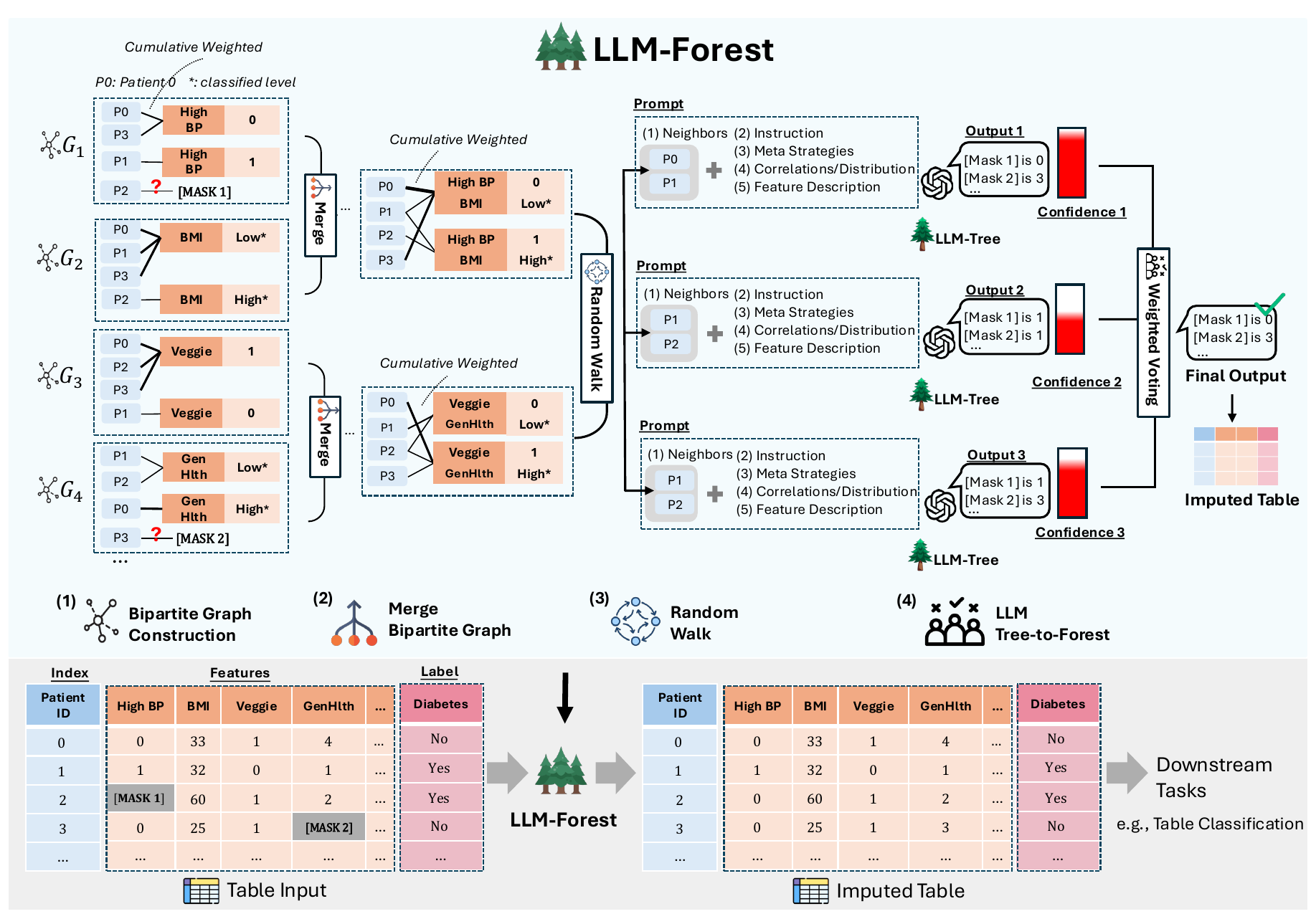}
  \caption{Overview of \sysn for data imputation. \sysn~ consists of four key steps: (1) Bipartite Information Graph construction: converts tabular data into bipartite information graphs, capturing the relationships between data entries and the values of their associated dimensional features. (2) Merge Bipartite Graphs: Hierarchically merges the bipartite information graphs to aggregate shared information across features. (3) Graph Retrieval via Random Walk: Performed on the merged graphs to identify correlated and diverse neighbors for each entry. (4) Construct LLM-tree to Forest: Builds a tailored few-shot prompt using the identified neighbors for each LLM; Multiple LLMs form a “forest” whose outputs are aggregated via confidence-weighted voting to generate the final imputed values.}
  \vspace{-0.20in}
  \label{fig_overview}
\end{figure*}

\section{Notations and Problem Definitions}
Let \([n] = \{1, 2, \dots, n\}\) and define \(N = [n]\). Suppose the matrix \(\mathbf{X} \in \mathbb{R}^{n \times d}\) represents a tabular dataset with \(n\) entries and \(d\)-dimensional features. The matrix \(\mathbf{X} = [A_1, A_2, \dots, A_n]^\top\), where each entry \(A_i \in \mathbb{R}^d\), \(i \in [n]\), is a \(d\)-dimensional vector. For example, in a diabetes dataset, \(A_i^\top\) could represent the feature set of the patient \(i\).
Let \(A_{ij}\), \(j \in [d]\), denote the \(j\)-th feature value of the \(i\)-th entry \(A_i^\top\). For convenience, we use \(\{C_1, C_2, \dots, C_d\}\) to represent the \(d\)-dimensional features (columns) of \(\mathbf{X}\), where \(C_{ji} \in \mathbb{R}^n\), \(j \in [d], i \in N\), denotes the \(i\)-th entry of the \(j\)-th feature. Let \(R_j\) represent the set of distinct values in feature \(C_j\), which can be viewed as the collection of attribute values for all patients in the dataset.
Additionally, define a matrix \(\mathbf{M} \in \mathbb{R}^{n \times d}\) to represent the missingness pattern of \(\mathbf{X}\), where a cell has the value 1 if the corresponding entry in \(\mathbf{X}\) is missing, and 0 otherwise.

The objective is to predict the missing values in \(\mathbf{X}\) based on the available information in the dataset.

\section{Proposed Method: LLM-Forest}

In this section, we present our proposed framework, \sysn, which consists of four key components, as illustrated in Figure~\ref{fig_overview}. The first three steps form the information-based graph retrieval module: (1) Table to Bipartite Information Graphs: This step converts tabular data into bipartite information graphs connecting dimensional features and their corresponding entries, where edges are weighted based on self-information to quantify the informativeness of feature values, enriching the representation of feature–entry relationships. (2) Hierarchical Bipartite Graph Merging: The bipartite graphs are then hierarchically merged, aggregating shared information across different features. (3) Graph Retrieval via Random Walk: Random walks are performed on the merged bipartite graphs to identify correlated and diverse neighbor nodes for each entry (node), preparing for the construction of the LLM-tree. Finally, (4) Construct LLM-tree to Forest: Based on the identified neighbors, a tailored few-shot prompt is constructed for each LLM, forming a forest with confidence-weighted voting. We detailed each procedure below.

\subsection{Table to Bipartite Information Graphs}

Given the dataset \(\mathbf{X}\) with \(n\) entries and \(d\)-dimensional features, we construct \(d\) bipartite graphs. For each \(j \in [d]\), a bipartite graph \(G_j = (N \cup R_j, E_j)\) is constructed, where \(E_j\) represents the set of edges connecting the two independent sets \(N\) (the set of entries) and \(R_j\) (the set of distinct values in feature \(j\) with continuous features discretized into bins).
For any \(i \in N\) and a distinct value \(a_k^j \in R_j\), an edge \(e_i(a_k^j)\) is formed if \(A_{ij} = a_k^j\). In this way, we construct \(d\) bipartite graphs \(\{G_1, G_2, \dots, G_d\}\), as illustrated on the left side of Figure \ref{fig_overview} (1).
Next, we will elaborate on the calculation of the weight \(w_i(a_k^j)\) for each edge \(e_i(a_k^j)\).

Given an entry \(A_i\), a dimension \(C_j\), and a value \(a_k^j \in R_j\), suppose that the probability of the event \(A_{ij} = a_k^j\) is \(p_{i}(a_k^j)\). Grounded in information theory \cite{ash2012information}, the self-information of the event that entry \(A_i\) takes the value \(a_k^j\) in dimension \(j\) is defined as:

\begin{equation}
w_{i}(a_k^j) = \log \left( 1 + p_{i}(a_k^j) \right).
\end{equation}
It is important to note that for an entry with a missing value in dimension \(j\), an isolated node will be formed in the corresponding bipartite graph \(G_j\). To address this challenge, we introduce a merging process, which will be detailed in Section \ref{sec:merge}.

\para{Determine $p_{i}(a^k_j)$}. To compute the informativeness-based edge weight \(w_i(a_k^j)\), we estimate the probability \(p_i(a_k^j)\) based on the type of feature \(C_j\). We consider three cases:

1. For a categorical feature \(C_j\), where \(j \in [d]\) is an attribute feature, we assume a uniform distribution and set:
\begin{equation}
    p_{i}(a_k^j) = \frac{1}{|R_j|},
\end{equation}
where \(R_j\) represents the set of distinct values of \(C_j\). This approximation works well for randomized features such as gender.

2. For features that follow a normal distribution, we approximate \( p_{i}(a_k^j) \) using the probability density function:
\begin{equation}
     p_{i}(a_k^j) = \frac{1}{\sigma_j \sqrt{2\pi}} \exp\left( -\frac{(a_k^j - \mu_j)^2}{2\sigma_j^2} \right),
\end{equation}
where \(\mu_j\) is the mean and \(\sigma_j^2\) is the variance of \(C_j\). This approximation is suitable for features that are expected to follow a normal distribution, such as weight or body mass index.

3. For features with skewed or unknown distributions, we estimate \( p_{i}(a_k^j) \) based on the empirical distribution:
\begin{equation}
    p_{i}(a_k^j) = \frac{|\{i \in N: A_{ij} =  a_k^j\}|}{n}.
\end{equation}
This method is effective for features with skewed distributions and low entropy, such as income or salary.

Therefore, we can construct \(d\) bipartite information graphs for the \(n\) entries with \(d\) features, where the edge weights in each graph reflect the information shared among entries through the feature value nodes. To effectively compress the information shared by entries across multiple features, we introduce the following approach to merge any two informative bipartite graphs.

\para{Time complexity}. 
We propose constructing \(d\) bipartite graphs to avoid building the conventional \(n \times n\) similarity graph, such as those used in KNN methods. In the conventional approach, calculating the similarity between two nodes (entries) takes \(O(d)\), resulting in a total complexity of \(O(n^2d)\) to construct the adjacency similarity matrix. In contrast, constructing each bipartite graph only requires two scans of the corresponding feature dimension, leading to a total complexity of \(O(nd)\) to build all \(d\) bipartite graphs. Consequently, this approach significantly reduces the computational cost of graph construction.

\vspace{-0.1cm}

\subsection{Hierarchical Bipartite Graph Merging} \label{sec:merge}

We use a bottom-up hierarchical process to merge the \(d\) bipartite graphs into compressed bipartite information graphs. In each step, two bipartite graphs are merged into one, while accumulating the shared information of entries across the two dimensional features. Given two bipartite graphs \(G_j = (N \cup R_j, E_j)\) and \(G_{j'}= (N \cup R_{j'}, E_{j'})\), where \(j, j' \in [d]\), we merge the right sides \(R_j\) and \(R_{j'}\), as both graphs share the same left side \(N\).

The goal is to merge two nodes \(a_k^j \in R_j\) and \(a_{k'}^{j'} \in R_{j'}\) if they have many shared neighbors in \(N\). Let \(\mathcal{N}(a_k^j)\) represent the set of nodes in \(N\) connected to \(a_k^j\) in \(G_j\), i.e., \(\mathcal{N}(a_k^j) = \{i \in N: A_{ij} = a_k^j \} \subseteq N\). We use the Jaccard coefficient to measure the similarity between two nodes:
\begin{equation}
S(a_k^j, a_{k'}^{j'}) = \frac{|\mathcal{N}(a_k^j) \cap \mathcal{N}(a_{k'}^{j'})|}{|\mathcal{N}(a_k^j) \cup \mathcal{N}(a_{k'}^{j'})|}.
\end{equation}
We set a threshold \(\sigma\) to determine whether to merge two value nodes. We merge \(a_k^j\) and \(a_{k'}^{j'}\) if \(S(a_k^j, a_{k'}^{j'}) \geq \sigma\). The merged node is denoted by \(a_{\hat{k}}^{\hat{j}} = \{a_k^j, a_{k'}^{j'}\}\). The edges are merged as follows:
\begin{equation}
\begin{aligned}
&\forall i \in \mathcal{N}(a_k^j) \cup \mathcal{N}(a_{k'}^{j'}), \\
&w_i(a_{\hat{k}}^{\hat{j}}) = w_i(a_k^j) + w_i(a_{k'}^{j'}).
\end{aligned}
\end{equation}

This rule ensures that the information shared by entry \(i\) is accumulated across both dimensions. Note that \(w_i(a_k^j) = 0\) if the edge \(e_i(a_k^j)\) does not exist.
By merging nodes in descending order of the Jaccard coefficient \(S\), a new bipartite graph \(G_{\hat{j}} = (N \cup R_{\hat{j}}, E_{\hat{j}})\) is formed.

Using this merging rule, we recursively apply the merging process, as illustrated in Figure \ref{fig_overview} (2). At each level of merging, we prioritize merging two nodes whose corresponding right-side sets are of similar size, i.e., \( |R_j| \approx |R_{j'}| \). This approach ensures a balanced merging process.

\para{Time complexity}.
Merging two bipartite graphs \(G_j\) and \(G_i\) takes \(O(n)\) time. To ensure high-quality merging, we can apply the Jaccard similarity measure, which requires \(O(|R_j| \times |R_{j'}| + n)\) time to compute for each pair of graphs. However, the merging process can be parallelized, allowing it to scale efficiently for large datasets. Consequently, the hierarchical merging process for \(d\) bipartite graphs requires \(O(\log_2(d))\) levels.
 
\subsection{Graph Retrieval via Random Walk}

After the recursive merging process, we obtain a set of merged bipartite graphs. Given a target node $\hat{i}$, for which we aim to predict the missing values, the next step is to identify neighboring nodes that share strong informational edges with it. These selected neighbors will then be used to construct the LLM-forest.

To achieve this, we propose using a simple yet effective approach: the random walk. In a bipartite graph $G_j$, the random walk begins from the target node $\hat{i}$. The transition matrix $\mathbf{P}^{j,1} \in \bbr^{n \times {|R_j|}}$, which represents transitions from $N$ to $R_j$, is defined as:
\begin{equation}
    \mathbf{P}_{ik}^{j,1} =
    \begin{cases}
\frac{w_i(a^j_k)}{ \sum_{a_{k'}^j \in \caln(i) } e^{w_i(a^j_{k'})}}, \ \text{if} \ (i, a_k^j) \in E_j \\
0, \ \text{otherwise}
    \end{cases}
\end{equation}
for $i \in N, a^j_k \in R_j$.
And the transition matrix $\mathbf{P}^{j,2} \in \bbr^{{|R_j|} \times n }$, which represents transitions from  $R_j$ to $N$, is defined as
\begin{equation}
    \mathbf{P}_{ki}^{j,2} =
    \begin{cases}
\frac{w_i(a^j_k)}{ \sum_{i' \in \caln(a^j_k) } e^{w_{i'}(a^j_{k})}}, \ \text{if} \ (i, a_k^j) \in E_j \\
0 \ \text{otherwise}
    \end{cases}
\end{equation}
for $a^j_k \in R_j, i \in N$.
The walker moves to the next node according to either $\mathbf{P}_1^j$ or $\mathbf{P}_2^j$. To identify $q$ neighboring nodes, we first perform $q$ rounds of random walks on each merged bipartite graph, selecting $q$ neighbors per graph. We then take the union of the nodes found across all merged graphs and assign each node a score, calculated as the average edge weight along its random walk path. Finally, we rank the nodes based on these scores and choose the top $q$ nodes. Since we are working with bipartite graphs, the number of steps for the random walk can be set to 2 or 4.

\subsection{Construct LLM-tree to Forest}

Inspired by ensemble learning, we introduce the LLM Forest, which consists of multiple LLM Trees. Each tree is designed as a diverse few-shot learner using an LLM. 
Specifically, each tree is prompted with a set of neighboring nodes returned by random walk on the compressed bipartite graphs. Suppose there are $m$ trees in \sysn, and then run the graph merging process $m$ times.
This aim is to provide high-quality and diverse information for each LLM. In addition to this, we carefully design the prompt for each LLM-tree as follows. Note that we employ a single LLM model, feeding the prompts sequentially and allowing the LLM to make the necessary inferences, to conserve resources.

\para{Promept design}.
To bridge the gap between the structured format of tabular data and the natural language processing capabilities of LLMs, our methodology incorporates a \textit{table transformation module}. This module utilizes a simple template to convert structured tabular data into a natural language format. For each entry (representing an individual sample), we extract the data across various features, such as age, gender, smoking status, and more from the structured dataset. These feature-value pairs are then formatted into a sample using the template: for each feature, if a value is present, it is included as \textit{"{feature}: {value};"} and we add a statement as "the sample has missing features: {missing feature 1, missing feature 2, ...}." at the end for all the missing values. 

Using the extracted neighbors' information for each entry, we structure the graph-augmented prompt to include several key components: the sample and their neighbors' data, meta strategies, dataset-specific correlations and distributions, dataset and feature descriptions, and instructions. We detail each component of the prompt in Appendix \ref{sec:prompt} in Table \ref{tab:prompt_example}.

\para{Confidence-weighted Voting}.
The most straightforward method for determining the ensemble output of the LLM Forest is majority voting. However, leveraging the self-assessment ability, we propose an enhanced strategy by weighting each LLM-tree’s vote according to its confidence level. Current LLMs can provide a confidence level associated with their inferences for specific tasks. For example, GPT4 assigns "High", "Medium" or "Low" confidence levels to its inference output. A higher confidence reflects a stronger alignment between the model's output and its internal logic and prior knowledge. Therefore, we introduce a criterion where higher confidence results in a greater voting weight. For instance, LLM trees with "High" confidence receive a voting weight of 1.0, while those with "Medium" confidence are assigned a weight of 0.6, "Low" with 0.3. The final imputed value is determined by selecting the value with the largest cumulative weighted votes.

\section{Experiments}
In this section, we evaluate \model~ across various real-world datasets under different settings. 

\subsection{Exepriment setup} 

We evaluate the imputation performance of the proposed method using 9 datasets from different domains: NPHA 
\cite{national_poll_on_healthy_aging_(npha)_936}, Gliomas 
\cite{glioma_grading_clinical_and_mutation_features_759}, Cancer
\cite{differentiated_thyroid_cancer_recurrence_915}, Credit-g \cite{statlog_(german_credit_data)_144}, Concrete \cite{concrete_compressive_strength_165}, Yacht \cite{yacht_hydrodynamics_243} and Wine \cite{wine_quality_186} datasets from the UCI Machine Learning Repository and Diabetes 
 and Housing \cite{harrison1978hedonic} dataset from Kaggle. The detailed information on datasets, the preprocessing procedures, the construction of the training and test sets, and the evaluation metrics are elaborated in Appendix \ref{sec:setting}.

We compare \model~with the baselines including both statistic and deep learning-based methods as described in Appendix \ref{sec:setting}. We also introduce zero-shot LLM (without similar samples provided): LLM-zero and adapt Chain-of-Thought \cite{wei2022chain} framework by providing example questions alongside corresponding imputation results and explanations for LLMs for comparison.
The method proposed in \cite{nazir2023chatgpt} generating a separate question for each missing value can be considered a variant of the LLM-zero.

Main experiments are conducted with GPT-4o (referred to as GPT-4 in the following discussion) and we also provided the results with Claude-3.5 and open-source model Mixtral-8×22B-v0.1 in the Appendix \ref{sec:claude}. The experiments are conducted 3 times to ensure reliability. The performance improvements are statistically significant with p-values less than 0.05 when compared to baselines. The hyperparameter settings for \model~and baseline methods are provided in Appendix \ref{sec:setting}.

\subsection{Main Results \& Discussion}

\para{Imputation Performance}.
We begin by evaluating the imputation accuracy of our proposed methods, as presented in Table~\ref{tab:maintable}. \model~achieves the strong imputation performance across all datasets when compared to traditional imputation techniques and LLM-based in-context learning baselines. From the main experiments, we derive the following insights:

\textit{(i) LLMs exhibit strong performance for data imputation across datasets with varying feature distributions}. For instance, In Gliomas, over 70\% of features are highly skewed, with over 90\% of values in one category. Despite this, \model~still outperforms statistical baselines.
\textit{(ii) LLMs are effective for imputing continuous datasets}. As shown in Table~\ref{tab:maintable}, \model~achieves the lowest MAE on most continuous datasets and ranks second on the remaining one. These results demonstrate the capability of LLMs to generate accurate values when faced with a wide range of possible outcomes in continuous imputation tasks.
\textit{(ii) A single LLM (LLM-Tree), when provided with an appropriately designed prompt, is capable of delivering accurate imputations.} Across all datasets, the LLM-Tree achieves performance within less than 1\% of the best method, consistently maintaining a high level of accuracy.
\textit{(iii) LLMs' own knowledge is limited, and they need to learn from informative contexts to perform well in this task.} In the zero-shot LLM setting, without informative neighbors, LLM-Zero leaves 9.27\% of Gliomas and 9.52\% of Diabetes cells unimputed. It also produces uncertain or unreasonable outputs, like "approximately" or "18000" for Age.
\textit{(iv) Forming an LLM forest enhances the accuracy and robustness of imputation results.} Compared to a single LLM, using just three trees can lead to performance improvements across all datasets. This highlights the effectiveness of ensemble learning in improving imputation robustness and accuracy of LLMs.

\begin{figure}[t]
  \centering
 
  \vspace{-0.05in}
  \includegraphics[width=\columnwidth]{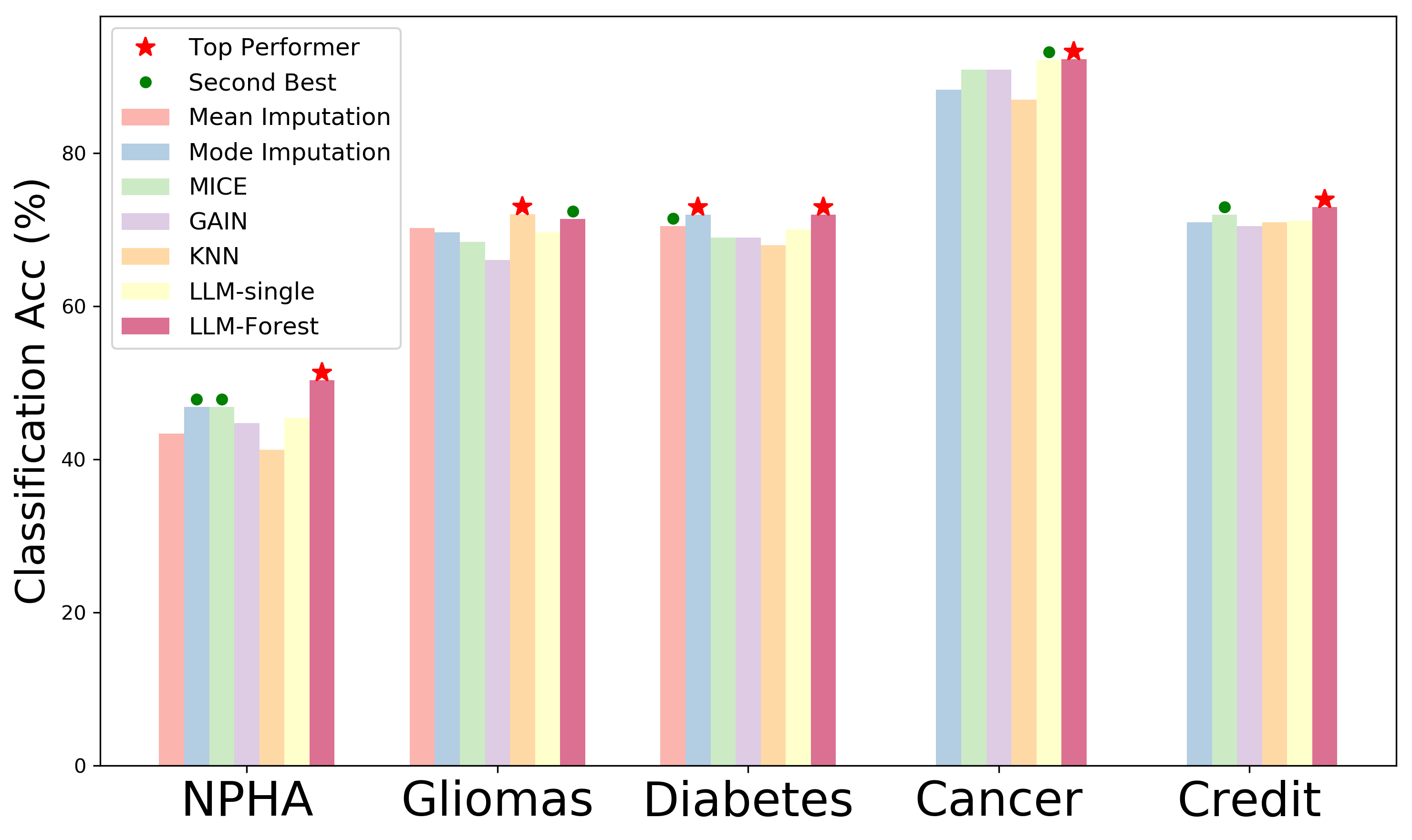}
  \caption{Performance on downstream classification task.}
  \vspace{-0.20in}
 \label{fig:lr}
\end{figure}

\para{Performance on Downstream tasks}.
We evaluate the downstream classification task performance using a logistic regression model \cite{cox1958regression} on five datasets. As shown in Figure \ref{fig:lr}, \model~consistently achieves the highest performance across 4 datasets and second-best on Gliomas. This superior performance can be attributed to the LLMs' ability to identify relevant patterns among features and entries during imputation, as well as its capacity to effectively apply its extensive base knowledge.

\begin{table*}[!th]
\centering
\small

\resizebox{\linewidth}{!}{
\begin{tabular}{l|ccccc|cccc}
\toprule
\multirow{2}{*}{\textbf{Methods}} & \textbf{NPHA} & \textbf{Gliomas}& \textbf{Diabetes} & \textbf{Cancer} &\textbf{Credit-g} & \textbf{Concrete} & \textbf{Yacht}& \textbf{Wine} & \textbf{Housing} \\
\cmidrule{2-10}
& ACC$\uparrow$ & ACC$\uparrow$ & ACC$\uparrow$ & ACC$\uparrow$ & ACC$\uparrow$ & MAE$\downarrow$ & MAE$\downarrow$ & MAE$\downarrow$ & MAE$\downarrow$ \\

\midrule
\textbf{\textit{Statistic \& Deep Learning}} \\

Mean Imputation &64.88 &83.29 &53.90 & -& - &  0.1814   &  0.2185 &  0.0988 & 0.1847 \\
Mode Imputation &\underline{66.10}&  83.29& 61.25 & 68.39 & 53.49 &  0.2568   & 0.2256  & 0.1142  &  0.2308 \\
MICE & 62.69& \underline{84.15}& 57.54& 42.52 & 44.00 & 0.1357    & 0.2002  &  0.0921 &  0.1115 \\
GAIN & 60.68 &84.13 & 54.18& 42.52& 53.49 &  0.1831   &   1.3642& 0.1085  &  0.1381 \\
KNN &64.28 &83.92 & 62.07 & \underline{71.98} &  \underline{54.15} &   0.1580  & 0.1937  & 0.1060  & 0.1130 \\
GRAPE &65.01 &81.36 & 61.26 & 68.90 &  53.48 & 0.1816   & 0.2150 &  0.1564 & 0.2493 \\

Miracle & 56.05 & 77.73 & 52.47 & 32.07 & 35.73 & 0.1320 & 0.2189 & 0.0820 & 0.1655 \\
Hyperimpute & 58.80 & 72.00 & 52.35 & 58.25 & 41.11 & 0.1310 & 0.1807 & 0.0788 & 0.1315 \\
TDM & 66.29 & 79.80 & 52.10 & 32.67 & 42.52 & 0.1678 & 0.1668 & 0.1000 & 0.1376 \\
Remasker & 65.38 & 83.45 & 57.16 & 42.52 & 36.26 & 0.1159 & 0.1819 & \underline{0.0780} & \textbf{0.0948} \\

\midrule
\textbf{\textit{LLM-based}} \\ 
Chain-of-Thought & 63.81& 81.83& 61.24 & 71.06 & 45.63 & 0.1991  &  0.1739  & 0.0990  & 0.1212 \\
LLM-zero & 59.18& 77.86& 55.80 & 59.32 & 36.12 &   0.1959  &  0.2231 & 0.1123  &  0.1226\\

\textbf{LLM-Tree (Ours)}  & 65.57&83.55 &\underline{62.46} &71.52 & 54.08 &  \underline{0.1105}   & \underline{ 0.1528} & 0.0816   &  0.1074\\

\textbf{LLM-Forest (Ours)} & \textbf{66.35}& \textbf{84.41}& \textbf{63.18}& \textbf{72.18} & \textbf{54.46} &  \textbf{0.1036}  & \textbf{0.1478}  & \textbf{0.0768}  &  \underline{0.1026}\\

\bottomrule
\end{tabular}
}
\caption{Comparison of imputation accuracy (ACC$\uparrow$) for categorical-dominant datasets (NPHA, Gliomas, Diabetes, Cancer, Credit-g) and mean absolute error (MAE$\downarrow$) for continuous datasets (Concrete, Yacht, Wine, Housing) with different baseline methods. ACC values are reported with \% omitted. The best results are highlighted in boldface, and the second-best results are underlined.}\label{tab:maintable}
\vspace{-0.3cm}
\end{table*}

\vspace{-0.1cm}
\subsection{Ablation Study}

\para{Impact of the different components in the prompt.} We conducted an ablation study on Meta Strategies (M-S) and Correlations and Distributions (C-D (Table~\ref{tab:component_study}). Meta Strategies aim to inspire the LLM by offering flexible reasoning approaches, leading to improved accuracy for NPHA (65.48\% $\rightarrow$ 66.35\%) and Gliomas (84.10\% $\rightarrow$ 84.41\%). Correlations and Distributions provide dataset-specific context, particularly for imbalanced features, and further boost accuracy for NPHA (65.73\% $\rightarrow$ 66.35\%) and Gliomas (83.55\% $\rightarrow$ 84.41\%). These demonstrate that M-S and C-D complement the core prompt components, enhancing the LLM's reasoning capabilities and contextual understanding.

\begin{table}[t]
\small

\centering
\scalebox{0.85}{
\begin{tabular}{l|cc|cc}
\toprule
\textbf{Datasets} & \textbf{w/ M-S} & \textbf{w/o M-S}  &\textbf{w/ C-D} & \textbf{w/o C-D} \\ 
\midrule

\textbf{NPHA}&66.35  &  65.48  &  66.35 & 65.73  \\
\textbf{Gliomas}& 84.41 & 84.10  & 84.41 &  83.55 \\

\bottomrule
\end{tabular}
}
\caption{Impact of the different components in the prompt. Results are with \% omitted. M-S: Meta Strategies, and C-D: Correlations and Distributions.}\label{tab:component_study}
\vspace{-0.6cm}
\end{table}

\para{Impact of tree count on LLM-Forest}. We analyze how the number of trees impacts the performance of the LLM-forest. Firstly, as we mentioned in the main results, \textit{the ensemble model outperforms a single LLM.} 
Secondly, 
while adding more trees allows us to provide LLMs with varied perspectives, the imputation results ultimately rely on the information present in the datasets and the LLM's external knowledge. Given that the pre-trained model for each few-shot LLM remains the same, thus, the information contained within the entire dataset and the external knowledge are both fixed, this poses an upper limit to the LLM’s inference capabilities. As shown in Table \ref{tab:treecount}, beyond three trees, the performance improvements plateau, with results such as only 0.1\% improvement for Cancer and 0.31\% for Gliomas when using 7 trees compared to using 3 trees. 
This highlights a trade-off between performance and computational cost. Although more trees can improve accuracy to some extent, the marginal benefits diminish and the additional inference steps introduce overhead. In practice, using a moderate number of trees (e.g., 3 to 5) offers a good balance between efficiency and performance.

\para{Effectiveness of confidence levels}. In LLM prompts, we include instructions for the models to output a confidence level for each imputed value, which we then use as weights when obtaining the ensemble results. We compute the imputation accuracy within different confidence level, the results are shown in Table \ref{tab:confidenceacc}.  High-confidence imputations consistently achieve higher accuracy across all datasets which proves that LLMs can evaluate the quality of their own outputs. 
This strong self-assessment capability plays a crucial role in improving the final outcomes by leveraging the strengths of each tree while downplaying less certain imputations. We further discuss the impact of confidence-weighted voting in Appendix \ref{sec:c-w}, and provide an additional sensitivity analysis of different confidence threshold settings in Appendix~\ref{sec:conf-thresh-sens}.

\begin{table}[t]
\small

\centering
\scalebox{0.8}{
\begin{tabular}{l|ccccc}
\toprule
\textbf{Confidence} &\textbf{NPHA} & \textbf{Gliomas} & \textbf{Diabetes} & \textbf{Cancer} \\ 
\midrule
\textbf{High} &  71.67 &  87.19 &  76.15   & 79.09\\
\textbf{Medium} & 51.42  & 63.15  &   40.18  & 38.79  \\ 
\textbf{Low} & 50.00  &  66.17 &   41.67  & 25.00 \\ 
\midrule
\textbf{M-V} & 65.91 &  84.35 & 63.33  & 72.08 \\ 
\textbf{C-W} & 66.35 &  84.41 & 63.18  & 72.18 \\ 


\bottomrule
\end{tabular}}

\caption{Imputation accuracy w.r.t. different confidence levels with \% omitted, with Majority Voting (M-V) and Confidence-Weighted (C-W) Aggregation.}\label{tab:confidenceacc}

\end{table}
\vspace{-0.2cm}

\begin{figure}[t]
  \centering
  
  \vspace{-0.05in}
  \includegraphics[width=0.7\columnwidth]{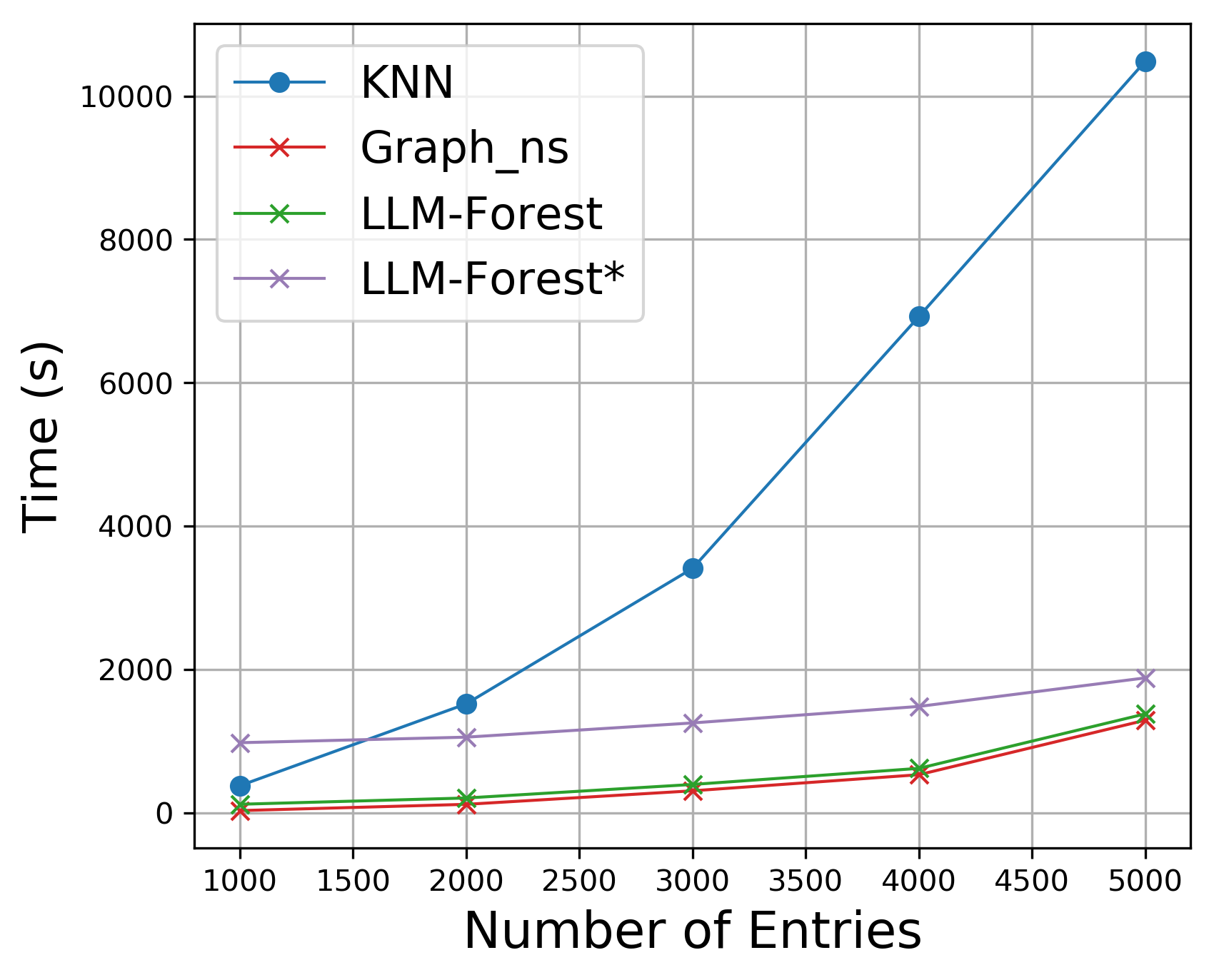}
  \caption{Time efficiency verification. LLM-Forest refers to processing 1 record per process in parallel, while LLM-Forest* refers to processing 10 records per process in parallel.}
  \vspace{-0.20in}
  \label{fig:time}
\end{figure}

\subsection{Case study}

\para{Scalability and time efficiency analysis.} To analyze the scalability and efficiency of \model, we randomly sample 1,000–5,000 entries from the Diabetes dataset and measure the runtime for neighbor searching and the entire pipeline. As shown in Figure~\ref{fig:time}, the runtime for neighbor searching with KNN grows sharply, reaching 10,000 seconds for 5,000 records, while our graph-based retrieval method (Graph\_ns) remains below 2,000 seconds, demonstrating superior scalability. Importantly, the constructed graph is shared across all trees in the ensemble, eliminating the need for repeated computation and further reducing overhead. For the entire pipeline, KNN’s exponential growth remains a bottleneck, whereas \model~scales efficiently. Its parallelizable architecture ensures computational efficiency for large datasets while maintaining accuracy and robustness in imputation results. We also provide results on a large-scale balanced dataset in Appendix~\ref{appendix:large-scale}, where \model~demonstrates great performance and scalability.

\begin{table}[t]

\small

\centering
\scalebox{0.8}{
\begin{tabular}{lccccc}
\toprule
\textbf{Tree count} &\textbf{3} & \textbf{4} & \textbf{5} & \textbf{6} & \textbf{7} \\ 
\midrule
\textbf{Cancer}&72.18 & 71.98  & 72.23  & 72.34 & 72.28 \\
\textbf{Gliomas}& 84.41& 84.51  &84.62& 84.60 & 84.72  \\ 


\bottomrule
\end{tabular}
}
\caption{Impact of tree count on \model~performance. Results are with \% omitted.}\label{tab:treecount}
\vspace{-0.5cm}
\end{table}


\para{Effects on Different Types of Missing Data.}
We explore how \model~ performs under various types of missing data mechanisms which we introduce in detail in Appendix \ref{sec:missing}. The results on Diabetes dataset (Table \ref{tab:missingness}), show that our approach consistently outperforms the baselines across all types of missing data. The largest improvement is observed with MNAR data, where information loss is more significant. In such cases, statistical methods struggle to accurately infer the real relationships between features when entire categories are missing for several features. However, LLMs, leveraging their external knowledge, retain the ability to infer missing categories by understanding the meaning of missing features and analyzing their relationships with existing data, allowing them to perform well even in more challenging scenarios.
\begin{table}[h]

\small

\centering
\scalebox{0.73}{
\begin{tabular}{cccccccc}
\toprule
\textbf{Dataset}  &\textbf{Mode} & \textbf{MICE} & \textbf{GAIN}  & \textbf{KNN} & \textbf{LLM-Tree} & \textbf{LLM-Forest}\\ 
\midrule

\textbf{MCAR}   &  61.25&57.54  & 54.18  & 62.07 &\underline{62.46}& \textbf{63.18}\\ 
\textbf{MAR}& \underline{63.06} & 60.75 & 57.70 &63.04  &63.03 & \textbf{64.48}\\ 
\textbf{MNAR} & 38.12 & 41.34   &  37.37 & 41.36 &\underline{49.97} & \textbf{51.25}\\
\bottomrule
    \end{tabular}}
\caption{Comparison of \model~ against baseline imputation methods for three missingness patterns on Diabetes dataset. Results are with \% omitted. }\label{tab:missingness}
\vspace{-0.3cm}
\end{table}

\vspace{-0.2cm}
\section{Related Work}
\paragraph{LLM for Tabular Data}
Recent researches have extended the use of LLMs beyond natural language tasks to structured formats like tabular data \cite{lu2024large, zhang2024survey, fang2024large,liu2025selfelicit}. Reasoning on tabular data with LLMs can be approached in two key areas. 
(1) Global Table Understanding: GPT models face challenges in processing large tables due to several reasons such as the semi-structure of the tabular data \cite{sui2023tap4llm}, the limited token capacity of LLMs \cite{chen2024tablerag}, and abstract or unseen information provided in table cells \cite{ye2023large}. These challenges prevent LLMs from fully capturing the table’s structure and content for downstream tasks. To tackle this, several studies such as TaPas \cite{herzig2020tapas} and Chain-of-Table \cite{wang2024chain} have proposed advanced table encoding and table-reasoning-chain approaches to help LLMs better understand and reason with tables.  
(2) Adaptation to Tabular Formats: LLMs, like GPT4, are primarily trained on textual data, limiting their adaptability and effectiveness when applied to the distinct nature of tabular datasets. Recent works \cite{chen2022program, cheng2022binding, sui2023tap4llm, jiang2023structgpt, lu2024chameleon,zou2025gtr} have proposed prompting and table pre-processing methods incorporating natural language techniques for improved analysis of tabular data. Notably, previous works have focused on downstream tabular reasoning tasks, such as QA and classification. The application of LLMs in tabular data imputation remains significant room for development.

\paragraph{Tabular Data Imputation} 
Tabular data imputation aims to fill in missing values in structured datasets. Simple statistical methods like mean, mode, and median imputation are widely used but often yield biased results. Methods like Multiple Imputation by Chained Equations (MICE) \cite{little2019statistical} and MissForest \cite{stekhoven2012missforest} improve accuracy by iteratively estimating missing values based on other observed features, but struggle with high-dimensional or imbalanced data. Recently, deep learning-based models such as GAIN (Generative Adversarial Imputation Networks) \cite{yoon2018gain} and DIFFIMPUTE \cite{wen2024diffimpute} have been developed to model complex data distributions, which require large and balanced datasets making them less practical for applications where data can be limited. Graphs have also been proved a powerful tool to model structural data \cite{zheng2024pyg, 10.1145/3583780.3615039,fu2024llms}. GRAPE \cite{you2020handling}, which encodes relationships between data points by graphs, offers an alternative by leveraging network structures to inform imputation. Leveraging the capabilities of LLMs, Anam \cite{nazir2023chatgpt} and CLAIM \cite{hayat2024claim} explored their use for imputation; however, both approaches require resource-intensive fine-tuning. Our work focuses the less explored area of table imputation with fine-tuning-free approach, marking a new direction within downstream tabular tasks.
\vspace{-0.1cm}
\section{Conclusion}
In this paper, we propose a general and scalable framework, \model, designed to enhance LLM-based reasoning through ensemble prompting. The framework combines multiple LLM trees employing diverse few-shot learning, forming an LLM forest to achieve robust and accurate imputations. While \model~is task-agnostic by design, this work focuses on its instantiation for data imputation.
Specifically, to extract high-quality contextual information for LLM trees, we developed a graph-based retrieval algorithm. To mitigate LLM bias and uncertainty, we introduce a confidence-weighted voting mechanism that aggregates outputs from all LLM trees based on their self-assessed confidence levels. Comprehensive experiments on 9 real-world datasets in various domains demonstrate that \model~effectively improves imputation accuracy while maintaining scalability and efficiency. The results highlight the potential of combining ensemble learning of LLMs with graph-based retrieval methods to tackle the complex problem of data imputation.

\vspace{-0.1cm}
\section*{Limitations}
The limitations of this paper are stated as follows:

\begin{itemize}[noitemsep, left=0pt]
\vspace{-0.3cm}
    \item In our experiments, we use GPT-4 and Claude-3.5 as backbone models via the OpenAI and Anthropic APIs and open-source model Mixtral-8×22B-v0.1. While LLM-Forest is compatible with other causal language models, performance may vary with different models \cite{achiam2023gpt,TheC3}. Using more static large language model, such as Llama-3-70B \citep{dubey2024llama}, could further mitigate this variability, but would require significant computing resources, which are often constrained.

    \item The potential benefits of integrating our method with fine-tuning techniques have not been investigated. Such combinations could reveal additional insights into task dimensions or enhance model performance.
    \vspace{-0.1cm}
\end{itemize}

\vspace{-0.3cm}
\section*{Ethics Statement}
This work fully adheres to the ACL Ethics Policy. To the best of our knowledge, no ethical concerns are associated with this paper. We use only existing open-source datasets, none of which contain private, personally identifiable information or offensive content.

\section*{Acknowledgements}
This work is supported by National Science Foundation under Award No. IIS-2117902. The views and conclusions are those of the authors and should not be interpreted as representing the official policies of the funding agencies or the government.


\bibliography{custom}

\begin{thebibliography}{71}
\expandafter\ifx\csname natexlab\endcsname\relax\def\natexlab#1{#1}\fi

\bibitem[{The()}]{TheC3}

\newblock \href {https://api.semanticscholar.org/CorpusID:268232499} {The claude 3 model family: Opus, sonnet, haiku}.

\bibitem[{nat(2017)}]{national_poll_on_healthy_aging_(npha)_936}
 2017.
\newblock {National Poll on Healthy Aging (NPHA)}.
\newblock UCI Machine Learning Repository.
\newblock {DOI}: https://doi.org/10.3886/ICPSR37305.v1.

\bibitem[{Abraham et~al.(2022)Abraham, Rahman, and Kaur}]{abraham2022tablequery}
Abhijith~Neil Abraham, Fariz Rahman, and Damanpreet Kaur. 2022.
\newblock Tablequery: Querying tabular data with natural language.
\newblock \emph{arXiv preprint arXiv:2202.00454}.

\bibitem[{Achiam et~al.(2023)Achiam, Adler, Agarwal, Ahmad, Akkaya, Aleman, Almeida, Altenschmidt, Altman, Anadkat et~al.}]{achiam2023gpt}
Josh Achiam, Steven Adler, Sandhini Agarwal, Lama Ahmad, Ilge Akkaya, Florencia~Leoni Aleman, Diogo Almeida, Janko Altenschmidt, Sam Altman, Shyamal Anadkat, et~al. 2023.
\newblock Gpt-4 technical report.
\newblock \emph{arXiv preprint arXiv:2303.08774}.

\bibitem[{Agrawal et~al.(2023)Agrawal, Hegselmann, Lang, Kim, and Sontag}]{agrawal2023large}
M~Agrawal, S~Hegselmann, H~Lang, Y~Kim, and D~Sontag. 2023.
\newblock Large language models are zero-shot clinical information extractors. arxiv, 2022.

\bibitem[{Ai et~al.(2025)Ai, Wei, Chen, Zeng, Zhao, Varatkar, Rouhani, Tang, Tong, and He}]{10.1145/3690624.3709196}
Mengting Ai, Tianxin Wei, Yifan Chen, Zhichen Zeng, Ritchie Zhao, Girish Varatkar, Bita~Darvish Rouhani, Xianfeng Tang, Hanghang Tong, and Jingrui He. 2025.
\newblock \href {https://doi.org/10.1145/3690624.3709196} {Resmoe: Space-efficient compression of mixture of experts llms via residual restoration}.
\newblock In \emph{Proceedings of the 31st ACM SIGKDD Conference on Knowledge Discovery and Data Mining V.1}, KDD '25, page 1–12, New York, NY, USA. Association for Computing Machinery.

\bibitem[{Ash(2012)}]{ash2012information}
Robert~B Ash. 2012.
\newblock \emph{Information theory}.
\newblock Courier Corporation.

\bibitem[{Batista et~al.(2002)Batista, Monard et~al.}]{batista2002study}
Gustavo~EAPA Batista, Maria~Carolina Monard, et~al. 2002.
\newblock A study of k-nearest neighbour as an imputation method.
\newblock \emph{His}, 87(251-260):48.

\bibitem[{Beltagy et~al.(2019)Beltagy, Lo, and Cohan}]{beltagy2019scibert}
Iz~Beltagy, Kyle Lo, and Arman Cohan. 2019.
\newblock Scibert: A pretrained language model for scientific text.
\newblock \emph{arXiv preprint arXiv:1903.10676}.

\bibitem[{Bernardini et~al.(2023)Bernardini, Doinychko, Romeo, Frontoni, and Amini}]{bernardini2023novel}
Michele Bernardini, Anastasiia Doinychko, Luca Romeo, Emanuele Frontoni, and Massih-Reza Amini. 2023.
\newblock A novel missing data imputation approach based on clinical conditional generative adversarial networks applied to ehr datasets.
\newblock \emph{Computers in Biology and Medicine}, 163:107188.

\bibitem[{Borzooei and Tarokhian(2023)}]{differentiated_thyroid_cancer_recurrence_915}
Shiva Borzooei and Aidin Tarokhian. 2023.
\newblock {Differentiated Thyroid Cancer Recurrence}.
\newblock UCI Machine Learning Repository.
\newblock {DOI}: https://doi.org/10.24432/C5632J.

\bibitem[{Brick and Kalton(1996)}]{brick1996handling}
J~Michael Brick and Graham Kalton. 1996.
\newblock Handling missing data in survey research.
\newblock \emph{Statistical methods in medical research}, 5(3):215--238.

\bibitem[{Brown(2020)}]{brown2020language}
Tom~B Brown. 2020.
\newblock Language models are few-shot learners.
\newblock \emph{arXiv preprint arXiv:2005.14165}.

\bibitem[{Chen et~al.(2024)Chen, Miculicich, Eisenschlos, Wang, Wang, Chen, Fujii, Lin, Lee, and Pfister}]{chen2024tablerag}
Si-An Chen, Lesly Miculicich, Julian~Martin Eisenschlos, Zifeng Wang, Zilong Wang, Yanfei Chen, Yasuhisa Fujii, Hsuan-Tien Lin, Chen-Yu Lee, and Tomas Pfister. 2024.
\newblock Tablerag: Million-token table understanding with language models.
\newblock \emph{arXiv preprint arXiv:2410.04739}.

\bibitem[{Chen et~al.(2022)Chen, Ma, Wang, and Cohen}]{chen2022program}
Wenhu Chen, Xueguang Ma, Xinyi Wang, and William~W Cohen. 2022.
\newblock Program of thoughts prompting: Disentangling computation from reasoning for numerical reasoning tasks.
\newblock \emph{arXiv preprint arXiv:2211.12588}.

\bibitem[{Cheng et~al.(2022)Cheng, Xie, Shi, Li, Nadkarni, Hu, Xiong, Radev, Ostendorf, Zettlemoyer et~al.}]{cheng2022binding}
Zhoujun Cheng, Tianbao Xie, Peng Shi, Chengzu Li, Rahul Nadkarni, Yushi Hu, Caiming Xiong, Dragomir Radev, Mari Ostendorf, Luke Zettlemoyer, et~al. 2022.
\newblock Binding language models in symbolic languages.
\newblock \emph{arXiv preprint arXiv:2210.02875}.

\bibitem[{Cortez and Reis(2009)}]{wine_quality_186}
Cerdeira A. Almeida F. Matos~T. Cortez, Paulo and J.~Reis. 2009.
\newblock {Wine Quality}.
\newblock UCI Machine Learning Repository.
\newblock {DOI}: https://doi.org/10.24432/C56S3T.

\bibitem[{Cox(1958)}]{cox1958regression}
David~R Cox. 1958.
\newblock The regression analysis of binary sequences.
\newblock \emph{Journal of the Royal Statistical Society Series B: Statistical Methodology}, 20(2):215--232.

\bibitem[{Devlin(2018)}]{devlin2018bert}
Jacob Devlin. 2018.
\newblock Bert: Pre-training of deep bidirectional transformers for language understanding.
\newblock \emph{arXiv preprint arXiv:1810.04805}.

\bibitem[{Du et~al.(2024)Du, Melis, and Wang}]{du2024remasker}
Tianyu Du, Luca~Melis Melis, and Ting Wang. 2024.
\newblock Remasker: Imputing tabular data with masked autoencoding.
\newblock In \emph{International Conference on Learning Representations (ICLR’24)}. International Conference on Learning Representations.

\bibitem[{Dubey et~al.(2024)Dubey, Jauhri, Pandey, Kadian, Al-Dahle, Letman, Mathur, Schelten, Yang, Fan et~al.}]{dubey2024llama}
Abhimanyu Dubey, Abhinav Jauhri, Abhinav Pandey, Abhishek Kadian, Ahmad Al-Dahle, Aiesha Letman, Akhil Mathur, Alan Schelten, Amy Yang, Angela Fan, et~al. 2024.
\newblock The llama 3 herd of models.
\newblock \emph{arXiv preprint arXiv:2407.21783}.

\bibitem[{Fang et~al.(2024)Fang, Xu, Tan, Zhang, Hu, Qi, Nickleach, Socolinsky, Sengamedu, Faloutsos et~al.}]{fang2024large}
Xi~Fang, Weijie Xu, Fiona~Anting Tan, Jiani Zhang, Ziqing Hu, Yanjun~Jane Qi, Scott Nickleach, Diego Socolinsky, Srinivasan Sengamedu, Christos Faloutsos, et~al. 2024.
\newblock Large language models (llms) on tabular data: Prediction, generation, and understanding-a survey.
\newblock \emph{URL https://arxiv. org/abs/2402.17944}.

\bibitem[{Fu et~al.(2024)Fu, Fang, Li, Tong, Torvik, and He}]{fu2024llms}
Dongqi Fu, Liri Fang, Zihao Li, Hanghang Tong, Vetle~I Torvik, and Jingrui He. 2024.
\newblock What do llms need to understand graphs: A survey of parametric representation of graphs.
\newblock \emph{arXiv preprint arXiv:2410.12126}.

\bibitem[{Gerritsma and Versluis(1981)}]{yacht_hydrodynamics_243}
Onnink~R. Gerritsma, J. and A.~Versluis. 1981.
\newblock {Yacht Hydrodynamics}.
\newblock UCI Machine Learning Repository.
\newblock {DOI}: https://doi.org/10.24432/C5XG7R.

\bibitem[{Gong et~al.(2020)Gong, Sun, Feng, Qin, Bi, Liu, and Liu}]{gong2020tablegpt}
Heng Gong, Yawei Sun, Xiaocheng Feng, Bing Qin, Wei Bi, Xiaojiang Liu, and Ting Liu. 2020.
\newblock Tablegpt: Few-shot table-to-text generation with table structure reconstruction and content matching.
\newblock In \emph{Proceedings of the 28th International Conference on Computational Linguistics}, pages 1978--1988.

\bibitem[{Gruver et~al.(2024)Gruver, Finzi, Qiu, and Wilson}]{gruver2024large}
Nate Gruver, Marc Finzi, Shikai Qiu, and Andrew~G Wilson. 2024.
\newblock Large language models are zero-shot time series forecasters.
\newblock \emph{Advances in Neural Information Processing Systems}, 36.

\bibitem[{Harrison~Jr and Rubinfeld(1978)}]{harrison1978hedonic}
David Harrison~Jr and Daniel~L Rubinfeld. 1978.
\newblock Hedonic housing prices and the demand for clean air.
\newblock \emph{Journal of environmental economics and management}, 5(1):81--102.

\bibitem[{Hayat and Hasan(2024)}]{hayat2024claim}
Ahatsham Hayat and Mohammad~Rashedul Hasan. 2024.
\newblock Claim your data: Enhancing imputation accuracy with contextual large language models.
\newblock \emph{arXiv preprint arXiv:2405.17712}.

\bibitem[{He et~al.(2023)He, Wei, and He}]{10.1145/3583780.3615039}
Xinrui He, Tianxin Wei, and Jingrui He. 2023.
\newblock \href {https://doi.org/10.1145/3583780.3615039} {Robust basket recommendation via noise-tolerated graph contrastive learning}.
\newblock In \emph{Proceedings of the 32nd ACM International Conference on Information and Knowledge Management}, CIKM '23, page 709–719, New York, NY, USA. Association for Computing Machinery.

\bibitem[{Hernandez et~al.(2022)Hernandez, Epelde, Alberdi, Cilla, and Rankin}]{hernandez2022synthetic}
Mikel Hernandez, Gorka Epelde, Ane Alberdi, Rodrigo Cilla, and Debbie Rankin. 2022.
\newblock Synthetic data generation for tabular health records: A systematic review.
\newblock \emph{Neurocomputing}, 493:28--45.

\bibitem[{Herzig et~al.(2020)Herzig, Nowak, M{\"u}ller, Piccinno, and Eisenschlos}]{herzig2020tapas}
Jonathan Herzig, Pawe{\l}~Krzysztof Nowak, Thomas M{\"u}ller, Francesco Piccinno, and Julian~Martin Eisenschlos. 2020.
\newblock Tapas: Weakly supervised table parsing via pre-training.
\newblock \emph{arXiv preprint arXiv:2004.02349}.

\bibitem[{Hofmann(1994)}]{statlog_(german_credit_data)_144}
Hans Hofmann. 1994.
\newblock {Statlog (German Credit Data)}.
\newblock UCI Machine Learning Repository.
\newblock {DOI}: https://doi.org/10.24432/C5NC77.

\bibitem[{Huang et~al.(2025)Huang, Ban, Fu, Li, Dai, Li, and Wang}]{huang2025adaptive}
Zixuan Huang, Yikun Ban, Lean Fu, Xiaojie Li, Zhongxiang Dai, Jianxin Li, and Deqing Wang. 2025.
\newblock Adaptive sample scheduling for direct preference optimization.
\newblock \emph{arXiv preprint arXiv:2506.17252}.

\bibitem[{J{\"a}ger et~al.(2021)J{\"a}ger, Allhorn, and Bie{\ss}mann}]{jager2021benchmark}
Sebastian J{\"a}ger, Arndt Allhorn, and Felix Bie{\ss}mann. 2021.
\newblock A benchmark for data imputation methods.
\newblock \emph{Frontiers in big Data}, 4:693674.

\bibitem[{Jarrett et~al.(2022)Jarrett, Cebere, Liu, Curth, and van~der Schaar}]{jarrett2022hyperimpute}
Daniel Jarrett, Bogdan~C Cebere, Tennison Liu, Alicia Curth, and Mihaela van~der Schaar. 2022.
\newblock Hyperimpute: Generalized iterative imputation with automatic model selection.
\newblock In \emph{International Conference on Machine Learning}, pages 9916--9937. PMLR.

\bibitem[{Jazayeri et~al.(2020)Jazayeri, Liang, and Yang}]{jazayeri2020imputation}
Ali Jazayeri, Ou~Stella Liang, and Christopher~C Yang. 2020.
\newblock Imputation of missing data in electronic health records based on patients’ similarities.
\newblock \emph{Journal of healthcare informatics research}, 4(3):295--307.

\bibitem[{Jiang et~al.(2023)Jiang, Zhou, Dong, Ye, Zhao, and Wen}]{jiang2023structgpt}
Jinhao Jiang, Kun Zhou, Zican Dong, Keming Ye, Wayne~Xin Zhao, and Ji-Rong Wen. 2023.
\newblock Structgpt: A general framework for large language model to reason over structured data.
\newblock \emph{arXiv preprint arXiv:2305.09645}.

\bibitem[{Kyono et~al.(2021)Kyono, Zhang, Bellot, and van~der Schaar}]{kyono2021miracle}
Trent Kyono, Yao Zhang, Alexis Bellot, and Mihaela van~der Schaar. 2021.
\newblock Miracle: Causally-aware imputation via learning missing data mechanisms.
\newblock \emph{Advances in Neural Information Processing Systems}, 34:23806--23817.

\bibitem[{Li et~al.(2024{\natexlab{a}})Li, Shetty, Kamath, Jaiswal, Jiang, Ding, and Kim}]{li2024cancergpt}
Tianhao Li, Sandesh Shetty, Advaith Kamath, Ajay Jaiswal, Xiaoqian Jiang, Ying Ding, and Yejin Kim. 2024{\natexlab{a}}.
\newblock Cancergpt for few shot drug pair synergy prediction using large pretrained language models.
\newblock \emph{NPJ Digital Medicine}, 7(1):40.

\bibitem[{Li et~al.(2025)Li, Wang, Sundaram, and Liu}]{li2025zero}
Yunzhe Li, Junting Wang, Hari Sundaram, and Zhining Liu. 2025.
\newblock A zero-shot generalization framework for llm-driven cross-domain sequential recommendation.
\newblock \emph{arXiv preprint arXiv:2501.19232}.

\bibitem[{Li et~al.(2024{\natexlab{b}})Li, Zheng, Jin, Fu, Jing, Ban, He, and Han}]{li2024can}
Zihao Li, Lecheng Zheng, Bowen Jin, Dongqi Fu, Baoyu Jing, Yikun Ban, Jingrui He, and Jiawei Han. 2024{\natexlab{b}}.
\newblock Can graph neural networks learn language with extremely weak text supervision?
\newblock \emph{arXiv preprint arXiv:2412.08174}.

\bibitem[{Little and Rubin(2019)}]{little2019statistical}
Roderick~JA Little and Donald~B Rubin. 2019.
\newblock \emph{Statistical analysis with missing data}, volume 793.
\newblock John Wiley \& Sons.

\bibitem[{Liu et~al.(2025)Liu, Amjad, Adkathimar, Wei, and Tong}]{liu2025selfelicit}
Zhining Liu, Rana~Ali Amjad, Ravinarayana Adkathimar, Tianxin Wei, and Hanghang Tong. 2025.
\newblock Selfelicit: Your language model secretly knows where is the relevant evidence.
\newblock \emph{arXiv preprint arXiv:2502.08767}.

\bibitem[{Lu et~al.(2024{\natexlab{a}})Lu, Peng, Cheng, Galley, Chang, Wu, Zhu, and Gao}]{lu2024chameleon}
Pan Lu, Baolin Peng, Hao Cheng, Michel Galley, Kai-Wei Chang, Ying~Nian Wu, Song-Chun Zhu, and Jianfeng Gao. 2024{\natexlab{a}}.
\newblock Chameleon: Plug-and-play compositional reasoning with large language models.
\newblock \emph{Advances in Neural Information Processing Systems}, 36.

\bibitem[{Lu et~al.(2024{\natexlab{b}})Lu, Zhang, Zhang, and Chen}]{lu2024large}
Weizheng Lu, Jiaming Zhang, Jing Zhang, and Yueguo Chen. 2024{\natexlab{b}}.
\newblock Large language model for table processing: A survey.
\newblock \emph{arXiv preprint arXiv:2402.05121}.

\bibitem[{Mattei and Frellsen(2019)}]{mattei2019miwae}
Pierre-Alexandre Mattei and Jes Frellsen. 2019.
\newblock Miwae: Deep generative modelling and imputation of incomplete data sets.
\newblock In \emph{International conference on machine learning}, pages 4413--4423. PMLR.

\bibitem[{Min et~al.(2022)Min, Lyu, Holtzman, Artetxe, Lewis, Hajishirzi, and Zettlemoyer}]{min2022rethinking}
Sewon Min, Xinxi Lyu, Ari Holtzman, Mikel Artetxe, Mike Lewis, Hannaneh Hajishirzi, and Luke Zettlemoyer. 2022.
\newblock Rethinking the role of demonstrations: What makes in-context learning work?
\newblock \emph{arXiv preprint arXiv:2202.12837}.

\bibitem[{Nazir et~al.(2023)Nazir, Cheeema, and Wang}]{nazir2023chatgpt}
Anam Nazir, Muhammad~Nadeem Cheeema, and Ze~Wang. 2023.
\newblock Chatgpt-based biological and psychological data imputation.
\newblock \emph{Meta-Radiology}, 1(3):100034.

\bibitem[{Safavi-Naini et~al.(2024)Safavi-Naini, Ali, Shahab, Shahhoseini, Savage, Rafiee, Samaan, Shabeeb, Ladak, Yang et~al.}]{safavi2024vision}
Seyed Amir~Ahmad Safavi-Naini, Shuhaib Ali, Omer Shahab, Zahra Shahhoseini, Thomas Savage, Sara Rafiee, Jamil~S Samaan, Reem~Al Shabeeb, Farah Ladak, Jamie~O Yang, et~al. 2024.
\newblock Vision-language and large language model performance in gastroenterology: Gpt, claude, llama, phi, mistral, gemma, and quantized models.
\newblock \emph{arXiv preprint arXiv:2409.00084}.

\bibitem[{Sankar et~al.(2021)Sankar, Wang, Krishnan, and Sundaram}]{10.1145/3460231.3474268}
Aravind Sankar, Junting Wang, Adit Krishnan, and Hari Sundaram. 2021.
\newblock \href {https://doi.org/10.1145/3460231.3474268} {Protocf: Prototypical collaborative filtering for few-shot recommendation}.
\newblock RecSys '21, page 166–175, New York, NY, USA. Association for Computing Machinery.

\bibitem[{Shwartz-Ziv and Armon(2022)}]{shwartz2022tabular}
Ravid Shwartz-Ziv and Amitai Armon. 2022.
\newblock Tabular data: Deep learning is not all you need.
\newblock \emph{Information Fusion}, 81:84--90.

\bibitem[{Stekhoven and B{\"u}hlmann(2012)}]{stekhoven2012missforest}
Daniel~J Stekhoven and Peter B{\"u}hlmann. 2012.
\newblock Missforest—non-parametric missing value imputation for mixed-type data.
\newblock \emph{Bioinformatics}, 28(1):112--118.

\bibitem[{Sterne et~al.(2009)Sterne, White, Carlin, Spratt, Royston, Kenward, Wood, and Carpenter}]{sterne2009multiple}
Jonathan~AC Sterne, Ian~R White, John~B Carlin, Michael Spratt, Patrick Royston, Michael~G Kenward, Angela~M Wood, and James~R Carpenter. 2009.
\newblock Multiple imputation for missing data in epidemiological and clinical research: potential and pitfalls.
\newblock \emph{Bmj}, 338.

\bibitem[{Sui et~al.(2023)Sui, Zou, Zhou, He, Du, Han, and Zhang}]{sui2023tap4llm}
Yuan Sui, Jiaru Zou, Mengyu Zhou, Xinyi He, Lun Du, Shi Han, and Dongmei Zhang. 2023.
\newblock Tap4llm: Table provider on sampling, augmenting, and packing semi-structured data for large language model reasoning.
\newblock \emph{arXiv preprint arXiv:2312.09039}.

\bibitem[{Tasci and Zhuge(2022)}]{glioma_grading_clinical_and_mutation_features_759}
Camphausen Kevin Krauze Andra~Valentina Tasci, Erdal and Ying Zhuge. 2022.
\newblock {Glioma Grading Clinical and Mutation Features}.
\newblock UCI Machine Learning Repository.
\newblock {DOI}: https://doi.org/10.24432/C5R62J.

\bibitem[{Thirunavukarasu et~al.(2023)Thirunavukarasu, Ting, Elangovan, Gutierrez, Tan, and Ting}]{thirunavukarasu2023large}
Arun~James Thirunavukarasu, Darren Shu~Jeng Ting, Kabilan Elangovan, Laura Gutierrez, Ting~Fang Tan, and Daniel Shu~Wei Ting. 2023.
\newblock Large language models in medicine.
\newblock \emph{Nature medicine}, 29(8):1930--1940.

\bibitem[{Troyanskaya et~al.(2001)Troyanskaya, Cantor, Sherlock, Brown, Hastie, Tibshirani, Botstein, and Altman}]{troyanskaya2001missing}
Olga Troyanskaya, Michael Cantor, Gavin Sherlock, Pat Brown, Trevor Hastie, Robert Tibshirani, David Botstein, and Russ~B Altman. 2001.
\newblock Missing value estimation methods for dna microarrays.
\newblock \emph{Bioinformatics}, 17(6):520--525.

\bibitem[{Wang et~al.(2024{\natexlab{a}})Wang, Rathi, and Sundaram}]{10.1145/3640457.3688145}
Junting Wang, Praneet Rathi, and Hari Sundaram. 2024{\natexlab{a}}.
\newblock \href {https://doi.org/10.1145/3640457.3688145} {A pre-trained zero-shot sequential recommendation framework via popularity dynamics}.
\newblock In \emph{Proceedings of the 18th ACM Conference on Recommender Systems}, RecSys '24, page 433–443, New York, NY, USA. Association for Computing Machinery.

\bibitem[{Wang et~al.(2024{\natexlab{b}})Wang, Zhang, Li, Eisenschlos, Perot, Wang, Miculicich, Fujii, Shang, Lee et~al.}]{wang2024chain}
Zilong Wang, Hao Zhang, Chun-Liang Li, Julian~Martin Eisenschlos, Vincent Perot, Zifeng Wang, Lesly Miculicich, Yasuhisa Fujii, Jingbo Shang, Chen-Yu Lee, et~al. 2024{\natexlab{b}}.
\newblock Chain-of-table: Evolving tables in the reasoning chain for table understanding.
\newblock \emph{arXiv preprint arXiv:2401.04398}.

\bibitem[{Wei et~al.(2022)Wei, Wang, Schuurmans, Bosma, Xia, Chi, Le, Zhou et~al.}]{wei2022chain}
Jason Wei, Xuezhi Wang, Dale Schuurmans, Maarten Bosma, Fei Xia, Ed~Chi, Quoc~V Le, Denny Zhou, et~al. 2022.
\newblock Chain-of-thought prompting elicits reasoning in large language models.
\newblock \emph{Advances in neural information processing systems}, 35:24824--24837.

\bibitem[{Wen et~al.(2024)Wen, Wang, Yi, Ke, and Shen}]{wen2024diffimpute}
Yizhu Wen, Yiwei Wang, Kai Yi, Jing Ke, and Yiqing Shen. 2024.
\newblock Diffimpute: Tabular data imputation with denoising diffusion probabilistic model.
\newblock In \emph{2024 IEEE International Conference on Multimedia and Expo (ICME)}, pages 1--6. IEEE.

\bibitem[{Ye et~al.(2023)Ye, Hui, Yang, Li, Huang, and Li}]{ye2023large}
Yunhu Ye, Binyuan Hui, Min Yang, Binhua Li, Fei Huang, and Yongbin Li. 2023.
\newblock Large language models are versatile decomposers: Decompose evidence and questions for table-based reasoning.
\newblock \emph{arXiv preprint arXiv:2301.13808}.

\bibitem[{Yeh(1998)}]{concrete_compressive_strength_165}
I-Cheng Yeh. 1998.
\newblock {Concrete Compressive Strength}.
\newblock UCI Machine Learning Repository.
\newblock {DOI}: https://doi.org/10.24432/C5PK67.

\bibitem[{Yin et~al.(2020)Yin, Neubig, Yih, and Riedel}]{yin2020tabert}
Pengcheng Yin, Graham Neubig, Wen-tau Yih, and Sebastian Riedel. 2020.
\newblock Tabert: Pretraining for joint understanding of textual and tabular data.
\newblock \emph{arXiv preprint arXiv:2005.08314}.

\bibitem[{Yoon et~al.(2018)Yoon, Jordon, and Schaar}]{yoon2018gain}
Jinsung Yoon, James Jordon, and Mihaela Schaar. 2018.
\newblock Gain: Missing data imputation using generative adversarial nets.
\newblock In \emph{International conference on machine learning}, pages 5689--5698. PMLR.

\bibitem[{You et~al.(2020)You, Ma, Ding, Kochenderfer, and Leskovec}]{you2020handling}
Jiaxuan You, Xiaobai Ma, Yi~Ding, Mykel~J Kochenderfer, and Jure Leskovec. 2020.
\newblock Handling missing data with graph representation learning.
\newblock \emph{Advances in Neural Information Processing Systems}, 33:19075--19087.

\bibitem[{Zhang et~al.(2024)Zhang, Wang, Dou, Zhu, and Che}]{zhang2024survey}
Xuanliang Zhang, Dingzirui Wang, Longxu Dou, Qingfu Zhu, and Wanxiang Che. 2024.
\newblock A survey of table reasoning with large language models.
\newblock \emph{arXiv preprint arXiv:2402.08259}.

\bibitem[{Zhao et~al.(2023)Zhao, Sun, Dezfouli, and Bonilla}]{zhao2023transformed}
He~Zhao, Ke~Sun, Amir Dezfouli, and Edwin~V Bonilla. 2023.
\newblock Transformed distribution matching for missing value imputation.
\newblock In \emph{International Conference on Machine Learning}, pages 42159--42186. PMLR.

\bibitem[{Zheng et~al.(2024)Zheng, Jing, Li, Zeng, Wei, Ai, He, Liu, Fu, You et~al.}]{zheng2024pyg}
Lecheng Zheng, Baoyu Jing, Zihao Li, Zhichen Zeng, Tianxin Wei, Mengting Ai, Xinrui He, Lihui Liu, Dongqi Fu, Jiaxuan You, et~al. 2024.
\newblock Pyg-ssl: A graph self-supervised learning toolkit.
\newblock \emph{arXiv preprint arXiv:2412.21151}.

\bibitem[{Zou et~al.(2025{\natexlab{a}})Zou, Ban, Li, Qi, Qiu, Yang, and He}]{zou2025transformer}
Jiaru Zou, Yikun Ban, Zihao Li, Yunzhe Qi, Ruizhong Qiu, Ling Yang, and Jingrui He. 2025{\natexlab{a}}.
\newblock Transformer copilot: Learning from the mistake log in llm fine-tuning.
\newblock \emph{arXiv preprint arXiv:2505.16270}.

\bibitem[{Zou et~al.(2025{\natexlab{b}})Zou, Fu, Chen, He, Li, Zhu, Han, and He}]{zou2025gtr}
Jiaru Zou, Dongqi Fu, Sirui Chen, Xinrui He, Zihao Li, Yada Zhu, Jiawei Han, and Jingrui He. 2025{\natexlab{b}}.
\newblock Gtr: Graph-table-rag for cross-table question answering.
\newblock \emph{arXiv preprint arXiv:2504.01346}.

\end{thebibliography}
\bibliographystyle{acl_natbib}

\newpage
\appendix

\section{Appendix}
\label{sec:appendix}

\begin{table*}[t]

\small

\centering
\begin{tabular}{c|p{4cm}|p{9.5cm}}
\toprule
\textbf{Role}  & 
\textbf{Components} & \textbf{Contents} \\ 
\midrule
\multirow{6}* {System} & \multirow{2}* {Setup} & Specify the task. For example, "You are a helpful assistant tasked with filling in the missing values for respondent 1 in a health-related telephone survey." \\ 
\cmidrule(lr){3-3}
& \multirow{4}* {Meta strategies} & Provide several strategies for inspiration. For example, "You have the flexibility to determine the best approach for each missing feature. Possible methods may include in no particular sequence (1) Using the mode of similar neighbors (2)…(3) Applying your knowledge in health domain (4)…".  \\ 
\midrule
\multirow{15}*{User}  
& \multirow{4}*{Correlations and distributions} & Provide dataset-specific characteristics. For example, "Here are some patterns observed based on the correlations: Diabetes\_binary is correlated with high blood pressure (0.40);…;…;…. Almost everyone has some form of healthcare (mean = 0.97);…". \\ 
\cmidrule(lr){3-3}
& \multirow{5}*{Dataset/feature descriptions} & Include the introduction of the dataset and feature descriptions. For example, "The data is from a health-related telephone survey that is collected annually by the CDC…. Below is detailed information about the respondent's feature description. It follows the format of 
<Feature Name>: <Indicated Value> = <Indicated Value Description> …".  \\ 
\cmidrule(lr){3-3}
& \multirow{4}*{Sample and neighbors' data} & Provide the records of the patient for imputation and their neighbors' records. For example, "the records of the similar patients for patient 1 are: Similar patient records 1 are HighBP:1,...,...;...Please infer the missing values in patient 1's records:....".  \\ 
\cmidrule(lr){3-3}
& \multirow{2}*{Instruction} &  Give the output format. For example, "Give the imputation results in a succinct JSON format with the following structure: "feature name": "inferred value"".\\ 
\bottomrule
\end{tabular}
\caption{LLM tree's prompt components and the examples.}\label{tab:prompt_example}
\end{table*}

\subsection{Prompts Example}
\label{sec:prompt}
We present a detailed explanation of the prompts used for our few-shot learning LLM tree. Table \ref{tab:prompt_example} outlines each component of the designed prompt, along with a concise example for each part for Diabetes dataset.

\subsection{Experiments Setting}
\label{sec:setting}
\para{Datasets construction.}
Our experiments are conducted on 9 real-world public datasets (NPHA, Gliomas, Diabetes, Credit-g, Concrete, Yacht, Wine, Housing and Cancer), and the statistics of the processed datasets are shown in Table \ref{tab:statistics}. These datasets vary in feature types: some consist of categorical features only, others have continuous features only, and a few include a mix of categorical and continuous features. For the Diabetes dataset, we randomly sample 1,000 records for experimentation. To simulate missing data scenarios, we randomly mask 40\% of the values in each feature. The datasets are then split into training and testing sets using an 80:20 ratio for downstream classification task evaluation. To prevent data leakage, imputation is performed only on the training data, and the trained classifier is then applied directly to the (still masked) test set.

\begin{table*}[t]
\small

\centering
\begin{tabular}{cccccccccc}
\toprule
\textbf{Datasets} &\textbf{NPHA} & \textbf{Gliomas} & \textbf{Diabetes} & \textbf{Cancer}  & \textbf{Credit-g} &\textbf{Concrete} & \textbf{Yacht} & \textbf{Wine} & \textbf{Housing}\\ 
\midrule
\textbf{\# of features}&14 & 23 & 22  & 16 & 20& 8 &7 & 11& 14\\
\textbf{\# of entries}&714 & 839  &1000& 383  &1000 & 1030 &308 & 1599& 506\\ 


\bottomrule
\end{tabular}
\caption{The datasets statistics.}\label{tab:statistics}
\end{table*}

\para{Baseline methods.}
We compare our method against widely used tabular data imputation techniques, including approaches such as mode imputation and mean imputation (for numerical datasets only), K-Nearest Neighbors (KNN) \cite{batista2002study}, Multiple Imputation by Chained Equations (MICE) \cite{little2019statistical} and Generative Adversarial Imputation Networks (GAIN) \cite{yoon2018gain}. We further include advanced methods such as GRAEPE \cite{you2020handling}, which treats the missing data imputation as graph representation learning task; MIRACLE \cite{kyono2021miracle}, which integrates causal regularization into iterative refinement; HyperImpute \cite{jarrett2022hyperimpute}, a hybrid imputer with automatic model selection; TDM \cite{zhao2023transformed}, a transformed distribution matching approach; and Remasker \cite{du2024remasker}, which adopts a masked autoencoding framework. Since Cancer and Credit-g consist predominantly of non-numerical features, traditional methods like MICE requiring numerical input, cannot be directly applied. For these datasets, we use one-hot encoding to transform non-numeric features into numerical representations before applying imputation. After imputation, we reverse the transformation back to the original feature format and calculate the accuracy of the restored data. Notably, we didn't include the comparison with CLAIM \cite{hayat2024claim} since it fine-tunes the LLM for downstream tasks by replacing missing cells with LLM-generated descriptions rather than providing actual imputation values.

\begin{table}[!h]
\centering
\resizebox{1\linewidth}{!}{
\begin{tabular}{lccccc|cccc}
\toprule
\multirow{2}{*}{\textbf{Methods}} & \textbf{NPHA} & \textbf{Gliomas}& \textbf{Diabetes} & \textbf{Cancer}  \\
\cmidrule{2-5}
& ACC($\uparrow$) & ACC($\uparrow$) & ACC($\uparrow$) & ACC($\uparrow$) \\

\midrule
\textbf{Statistic $\&$ Deep Learning} \\
Mean Imputation &64.88 &83.29 &53.90 & - \\
Mode Imputation &66.10&  83.29& 61.25 & 68.39 \\
MICE & 62.69& 84.15& 57.54& 42.52  \\
GAIN & 60.68 &84.13 & 54.18& 42.52 \\
KNN &64.28 &83.92 & 62.07  & 71.98 \\
GRAPE &65.01 &81.36 & 61.26 & 68.90 \\
Miracle & 56.05 & 77.73 & 52.47 & 32.07  \\
Hyperimpute & 58.80 & 72.00 & 52.35 & 58.25 \\
TDM & 66.29 & 79.80 & 52.10 & 32.67  \\
Remasker & 65.38 & 83.45 & 57.16 & 42.52  \\
\midrule
\textbf{LLM-based} \\
Chain-of-Thought & 63.81& 81.83& 61.24 & 71.06  \\
LLM-zero & 59.18& 77.86& 55.80 & 59.32  \\
LLM-tree & \underline{65.57}&83.55 &62.46 &71.52 \\
LLM-Forest (Claude-3.5-based) & 64.41& 82.97& 59.02& 
\textbf{73.51}\\

LLM-Forest (Mixtral-based) & 63.91& \underline{84.34}  & \textbf{63.21}& 71.47 \\
LLM-Forest (GPT-4-based) & \textbf{66.35}& \textbf{84.41}& \underline{63.18}& \underline{72.18} \\		

\bottomrule
\end{tabular}}
\caption{Comparison of imputation accuracy of Mixtral-based \model~with different baseline methods on four datasets with $\%$ omitted. The best results are highlighted in boldface. Underlined values indicate the second best.} \label{tab:claude}
\end{table}

\para{Hyperparameter settings.}
We provide the experimental setting for \model~and baseline methods. For MICE, we applied the default settings as provided in the PyPI package. For the KNN approach, we selected k = 5 for Gliomas dataset, k = 7 for Daibetes, Cancer and NPHA datasets and k=10 for the rest, as the neighbors we used in \model~for fair comparison. For all other baselines, we adopt their original implementations and follow the hyperparameters recommended in their respective papers.

In the experiments, we set the number of trees in the LLM forest as 3. The threshold $\sigma$ for merging nodes on two bipartite graphs is set to 20 for all datasets. The jump steps for random walk on merged bipartite graphs are 2 in the experiments. The graph merging step is 1 for Gliomas and Wine dataset, 2 for Concrete, NPHA and Housing dataset, and 3 for the rest. 

\para{Evaluation metric.} We use two evaluation metrics to assess the imputation performance depending on the dataset type:  

For datasets primarily composed of \textbf{categorical features} including NPHA, Gliomas, Diabetes, Cancer and Credit-g, Imputation Accuracy (ACC) is computed based on exact matches between the imputed values and the ground truth. For those numerical features in the categorical datasets, we introduce a small tolerance to account for minor deviations, such as NPHA, Gliomas, and Diabetes, a tolerance of \(\pm 1\) is used for features like \textit{Age} and \textit{BMI}. On Credit-g dataset, for continuous feature \textit{credit amount}, accuracy is evaluated by matching the most significant digit.  

For datasets with predominantly \textbf{continuous features}, we compute the Mean Absolute Error (MAE) after performing min-max normalization. The MAE is then defined as:  
\begin{equation}
\text{MAE} = \frac{1}{N} \sum_{i=1}^N |x_i - \hat{x}_i|,
\end{equation}
where \(x_i\) is the ground truth value, \(\hat{x}_i\) is the imputed value, and \(N\) is the total number of imputed entries. This ensures a fair comparison across features with different ranges. MAE is applied to Concrete, Yacht, Wine, and Housing datasets.

\begin{table*}[h]
\small

\centering

\resizebox{0.85\linewidth}{!}{
\begin{tabular}{l|ccccccccc}
\toprule
\textbf{Dataset}&\textbf{Mask\%}&\textbf{n} &\textbf{Methods} &\textbf{Tree 1} & \textbf{Tree 2} & \textbf{Tree 3} & \textbf{Majority voting} & \textbf{Confidence-weighted} \\

\midrule
\multirow{2}{*}{\textbf{Gliomas}}
&\multirow{2}{*}{\textbf{40\%}} &\multirow{2}{*}{\textbf{5}}& Mode Impu  & 83.65 & 83.27 & 83.50 & 84.36 & - \\ 
&&&GPT-4 Impu & 83.53 & 83.55 & 83.35 & 84.35 & 84.41 \\ 
\midrule
\multirow{2}{*}{\textbf{Diabetes}}
&\multirow{2}{*}{\textbf{40\%}} &\multirow{2}{*}{\textbf{7}}& Mode Impu  & 60.61 & 59.33 & 59.87 & 61.82 & - \\ 
&&&GPT-4 Impu & 62.46 & 61.29 & 62.38 & 63.33 & 63.18 \\

\midrule
\multirow{2}{*}{\textbf{Cancer}}
&\multirow{2}{*}{\textbf{40\%}}&\multirow{2}{*}{\textbf{7}} & Mode Impu  & 70.08 & 70.34 & 70.39 &71.31 & - \\ 
&&&GPT-4 Impu & 71.11 & 71.11 & 71.52&  72.08& 72.18 \\ 

\midrule
\multirow{2}{*}{\textbf{NPHA}}

&\multirow{2}{*}{\textbf{40\%}}&\multirow{2}{*}{\textbf{7}} & Mode Impu & 64.54 & 64.51 & 65.41 & 65.80 & - \\ 
&&&GPT-4 Impu & 65.16 & 64.13 & 65.57 & 65.91 & 66.35 \\ 
\midrule
\multirow{2}{*}{\textbf{Credit-g}}

&\multirow{2}{*}{\textbf{40\%}}&\multirow{2}{*}{\textbf{7}} & Mode Impu & 52.13 & 51.64 & 53.09 & 53.23 & - \\ 
&&&GPT-4 Impu & 53.66 & 53.00 & 54.08 & 54.17 & 54.46 \\ 
\bottomrule

\end{tabular}
}
\caption{Imputation ccuracy (ACC$\uparrow$) on categorical-dominant datasets for individual LLM trees, majority voting, and confidence-weighted voting. The mask ratio is 40\%, and n represents the number of neighbors provided in the prompts. The results are reported with \% omitted.}\label{tab:trees}
\end{table*}
\begin{table*}[h!]
\small

\centering

\resizebox{0.85\linewidth}{!}{
\begin{tabular}{l|ccccccccc}
\toprule
\textbf{Dataset}&\textbf{Mask\%}&\textbf{n} &\textbf{Methods} &\textbf{Tree 1} & \textbf{Tree 2} & \textbf{Tree 3} & \textbf{Majority voting} & \textbf{Confidence-weighted} \\ 

\midrule
\multirow{2}{*}{\textbf{Concrete}}
&\multirow{2}{*}{\textbf{40\%}}&\multirow{2}{*}{\textbf{10}} & Mode Impu  & 0.1399 & 0.1391 & 0.1392  &0.1361 & - \\ 
&&&GPT-4 Impu & 0.1105 & 0.1108 & 0.1087 &  0.1085& 0.1036 \\ 
\midrule
\multirow{2}{*}{\textbf{Yacht}}
&\multirow{2}{*}{\textbf{40\%}}&\multirow{2}{*}{\textbf{10}} & Mode Impu  & 0.1988 & 0.1989 & 0.2094 & 0.1912 & - \\ 
&&&GPT-4 Impu & 0.1544 & 0.1528 & 0.1674&  0.1531& 0.1478 \\ 
\midrule
\multirow{2}{*}{\textbf{Wine}}
&\multirow{2}{*}{\textbf{40\%}}&\multirow{2}{*}{\textbf{10}} & Mode Impu  & 0.0870 & 0.0876 & 0.0877 & 0.0855 & - \\ 
&&&GPT-4 Impu & 0.0817 & 0.0816 & 0.0819&  0.0782& 0.0768 \\ 
\midrule
\multirow{2}{*}{\textbf{Housing}}
&\multirow{2}{*}{\textbf{40\%}}&\multirow{2}{*}{\textbf{10}} & Mode Impu  & 0.1388 & 0.1396 & 0.1395 & 0.1371& - \\ 
&&&GPT-4 Impu & 0.1074 & 0.1087 & 0.1087&  0.1059& 0.1026 \\ 

\bottomrule
\end{tabular}
}
\caption{Mean absolute error (MAE$\downarrow$) on continuous datasets for individual LLM trees, majority voting, and confidence-weighted voting. The mask ratio is 40\% and n is the number of neighbors provided in the prompts.}\label{tab:trees2}
\end{table*}

\subsection{Evaluating LLM-Forest with Claude-3.5 and Mixtral}
\label{sec:claude}

In addition to GPT-4, we evaluate the model~framework using Claude-3.5 and the open-source Mixtral (Mixtral-8×22B-v0.1), which aims to demonstrate the performance of the LLM-Forest framework when different base LLMs are utilized. As shown in Table \ref{tab:claude}, due to inherent performance differences between models \cite{safavi2024vision}, Claude-3.5 and Mixtral does not always achieve the top results compared to GPT-4, yet it consistently delivers competitive performance across most datasets.

Mixtral, which institutions can deploy locally, offers a privacy-preserving alternative to closed models. Despite its smaller size and lower general capabilities compared to GPT-4, Mixtral achieves the best result on Diabetes (63.21\%) and second-best on Gliomas (84.34\%), outperforming strong baselines like KNN and GRAPE on most datasets.

The Claude-3.5-based \model~achieves the best result on Cancer (73.51\%), demonstrating its strengths in this domain while underperforming on Diabetes dataset, likely due to limitations in its internal knowledge base regarding this specific domain \cite{safavi2024vision}.

Overall, these results showcase the adaptability of the \model~to both open-source and closed-source models, with Mixtral demonstrating strong potential for privacy-sensitive applications and Claude-3.5 proving to be a reliable and effective option for data imputation in specific domains.

\subsection{Effectiveness of Graph-based Retrieval and Confidence-weighted Voting}
\label{sec:c-w}
As shown in Tables \ref{tab:trees} and \ref{tab:trees2}, Mode Impu, which imputes missing values using the mode of neighbors retrieved by our graph-based retrieval algorithm, highlights the high quality of the retrieved neighbors, achieving competitive results, such as 65.80\% on NPHA and 84.36\% on Gliomas. Further, GPT-4 Impu leverages these high-quality neighbors as contextual input and consistently achieves better performance by identifying patterns and producing informed predictions. These results demonstrate that incorporating confidence levels allows the model to prioritize higher-quality predictions, leading to more accurate imputations, while Mode Imputation validates the effectiveness of the retrieved neighbors as a strong foundation for imputation.

\subsection{Confidence Threshold Sensitivity}
\label{sec:conf-thresh-sens}
We conducted a sensitivity analysis on the Cancer dataset using 7 trees, evaluating several alternative confidence threshold settings. Our default scheme assigns weights of 1.0, 0.6, and 0.3 to “High,” “Medium,” and “Low” confidence predictions, respectively, based on the straightforward principle that more confident predictions should exert greater influence.

As shown in Table~\ref{tab:conf_threshold_comparison}, the overall performance remains stable across reasonable weight configurations, with only minor variations. A key reason for this stability is that high-confidence predictions consistently dominate the results, as shown in Table \ref{tab:conf_distribution}. However, assigning disproportionately high weights to low-confidence predictions (e.g., [0.1, 1.0, 0.6]) results in performance degradation, sometimes even worse than using a single tree. This suggests that amplifying the influence of unreliable outputs can negatively affect the effectiveness of the ensemble.

\begin{table}[h!]
\small
\centering
\resizebox{\linewidth}{!}{
\begin{tabular}{lcccc}
\toprule
\makecell{\textbf{Confidence threshold}\\\textbf{[High, Medium, Low]}} 
& [1.0,0.6,0.3]& [1.0,0.4,0.2] & [1.0,0.3,0.2] & [0.1,1.0,0.6] \\
\midrule
\textbf{Cancer} & 71.93 & 71.57 & 71.47 & 70.80 \\
\bottomrule
\end{tabular}}
\caption{Performance under different confidence threshold combinations (Cancer dataset). Results are reported with \% omitted.}
\label{tab:conf_threshold_comparison}
\vspace{-0.4cm}
\end{table}

\begin{table}[h]
\centering
\small
\resizebox{\linewidth}{!}{
\begin{tabular}{l|ccccccc}
\toprule
\textbf{Confidence level} & \textbf{tree1} & \textbf{tree2} & \textbf{tree3} & \textbf{tree4} & \textbf{tree5} & \textbf{tree6} & \textbf{tree7} \\
\midrule
\textbf{High}   & 79.34 & 78.21 & 78.99 & 81.05 & 78.79 & 80.25 & 79.41 \\
\textbf{Medium} & 17.99 & 18.35 & 18.25 & 16.85 & 19.31 & 16.56 & 18.24 \\
\textbf{Low}    &  2.67 &  3.44 &  2.77 &  2.10 &  1.90 &  3.19 &  2.35 \\
\bottomrule
\end{tabular}}
\caption{Confidence level distribution across trees (with \% omitted)}
\label{tab:conf_distribution}
\vspace{-0.4cm}
\end{table}

Table \ref{tab:conf_distribution} further confirms that across different trees, our results are highly consistent, with "High" confidence consistently dominating (~80\%). This consistency indicates that regardless of specific values within a reasonable range, the overall decision-making trend remains unchanged and reliable.

\subsection{Results on Large Scale Dataset}
\label{appendix:large-scale}
To evaluate the performance of \model~on general large-scale balanced datasets, we constructed a balanced version of the Diabetes dataset 
 Diabetes\_L by sampling records according to the size of the smallest class, resulting in 56,552 records. Retrieval was conducted across the entire dataset, with performance evaluated on 600 entries to ensure efficient analysis within resource constraints. As shown in Table \ref{tab:large}, \model~achieves the best performance with an accuracy of 61.61\%. In this setting, KNN was not applicable due to its computational limitations and inability to scale to datasets of this size.
 
\begin{table}[h]

\small

\centering
\scalebox{0.73}{
\begin{tabular}{lcccccc}
\toprule
\textbf{Dataset}  & \textbf{Mode} & \textbf{MICE} & \textbf{GAIN}  & \textbf{KNN} & \textbf{GRAPE} & \textbf{LLM-Forest}\\ 
\midrule
\textbf{Diabetes\_L} & 61.54 & 59.81 & 54.99 & OTT & 61.02 & \textbf{61.61} \\
\bottomrule
\end{tabular}}
\caption{Performance on the large-scale balanced Diabetes dataset. OTT stands for "Over Time Threshold." Results are reported with \% omitted.}\label{tab:large}
\end{table}

\subsection{Effects on Different Types of Missing Data}
\label{sec:missing}

To create datasets with missingness, we introduced up to 40\% missing values using three distinct missingness mechanisms: MCAR (40\% masked), MAR (20\% masked), and MNAR (33\% masked). These methods were implemented following \cite{hayat2024claim, jager2021benchmark} customized for our experimental needs, starting from fully complete datasets.
\begin{itemize}[noitemsep, left=0pt]
    \item \textbf{Missing Completely at Random (MCAR)} In this scenario, missing values occur without any pattern as what we used in the above evaluations.
    \item \textbf{Missing at Random (MAR)}  The likelihood of data being missing is related to other observed variables. Following the setting in \cite{hayat2024claim}, we identify the observations within the 30th percentile of the label column and randomly remove 40\% of the values in the other columns for these selected observations.
    \item \textbf{Missing Not at Random (MNAR)} The missing data is related to the value of the missing variable itself. For binary features we apply self-dependent (make '1' values missing with 30\% probability) and cross-feature MNAR masking (make current value missing with 40\% probability if the next feature's value is '0' ). For other features, we remove observations that fall within the 30th percentile of the feature value.

\end{itemize}

\end{document}